\documentclass[10pt,twocolumn,letterpaper]{article}

\usepackage{cvpr}      
\definecolor{cvprblue}{rgb}{0.21,0.49,0.74}
\usepackage[pagebackref,breaklinks,colorlinks,allcolors=cvprblue]{hyperref}

\usepackage{amsmath,amssymb,amsfonts,bm,booktabs,tabularx,bbm,mathrsfs}
\usepackage{tabularx}
\usepackage[linesnumbered,ruled,vlined]{algorithm2e}
\usepackage[dvipsnames]{xcolor}
\definecolor{mygray}{HTML}{3D5C6F}
\definecolor{myred}{HTML}{E47159}
\definecolor{myyellow}{HTML}{F9AE78}

\def\paperID{***} 
\def\confName{CVPR}
\def\confYear{2026}

\title{PvP: Data-Efficient Humanoid Robot Learning with Proprioceptive-Privileged Contrastive Representations}

\author{Mingqi Yuan\textsuperscript{12}\thanks{Work done at LimX Dynamics.}, Tao Yu\textsuperscript{2}\thanks{Project lead}, Haolin Song\textsuperscript{24}, Bo Li\textsuperscript{1}, Xin Jin\textsuperscript{36}, Hua Chen\textsuperscript{25\textdaggerdbl}, Wenjun Zeng\textsuperscript{3}\thanks{Corresponding author}\\
\textsuperscript{1}HK PolyU$\quad$\textsuperscript{2}LimX Dynamics$\quad$\textsuperscript{3}EIT, Ningbo$\quad$\textsuperscript{4}USTC$\quad$\textsuperscript{5}ZJU-UIUC$\quad$\textsuperscript{6}ZGCA\\
{\tt\small mingqi.yuan@connect.polyu.hk, huachen@intl.zju.edu.cn, wzeng-vp@eitech.edu.cn}
}

\begin{document}

\maketitle

\begin{abstract}
Achieving efficient and robust whole-body control (WBC) is essential for enabling humanoid robots to perform complex tasks in dynamic environments. Despite the success of reinforcement learning (RL) in this domain, its sample inefficiency remains a significant challenge due to the intricate dynamics and partial observability of humanoid robots. To address this limitation, we propose \textbf{PvP}, a \textbf{P}roprioceptive-\textbf{P}rivileged contrastive learning framework that leverages the intrinsic complementarity between proprioceptive and privileged states. PvP learns compact and task-relevant latent representations without requiring hand-crafted data augmentations, enabling faster and more stable policy learning. To support systematic evaluation, we develop \textbf{SRL4Humanoid}, the first unified and modular framework that provides high-quality implementations of representative state representation learning (SRL) methods for humanoid robot learning. Extensive experiments on the LimX Oli robot across velocity tracking and motion imitation tasks demonstrate that PvP significantly improves sample efficiency and final performance compared to baseline SRL methods. Our study further provides practical insights into integrating SRL with RL for humanoid WBC, offering valuable guidance for data-efficient humanoid robot learning. Our code is available at the GitHub repository\footnote{\url{https://github.com/myismyname/SRL4Humanoid}}.
\end{abstract}

\section{Introduction}
Humanoid robots have emerged as a critical platform for embodied intelligence, whose human-like morphology provides inherent advantages such as versatility, adaptability to human-centered environments, and intuitive interaction \cite{darvish2023teleoperation,tong2024advancements,Valenzuela2024EmbodyingIH}. To enable these capabilities, whole-body control (WBC) is essential for coordinating numerous joints and actuators to achieve balanced, agile, and safe behaviors in real-world settings \cite{sentis2006whole,cheng2024expressive}. However, designing effective WBC policies is particularly challenging due to the complex dynamics of humanoids, underactuation, and strong coupling between locomotion, manipulation, and balance \cite{kuindersma2016optimization,yuan2025behavior}. As a result, traditional model-based methods often fail to ensure flexible real-time control and robust performance under non-stationary conditions \cite{sentis2005synthesis,romualdi2022online}.

To overcome these limitations, recent research has increasingly turned to data-driven approaches, with reinforcement learning (RL) emerging as a dominant paradigm for humanoid WBC \cite{liao2025beyondmimic,yuan2025behavior,xue2025unified,ze2025twist,huang2025learning,li2025clone}. For instance, BeyondMimic \cite{liao2025beyondmimic} leverages large-scale RL-based motion tracking to achieve strong generalization across diverse human-like whole-body motions, while HugWBC \cite{xue2025unified} employs RL to train a single policy for multi-gait locomotion (\textit{e.g.}, standing, walking, and jumping) within a unified command space. Despite these promising results, RL applications for humanoid WBC remain constrained by sample inefficiency. The intricate dynamics and partial observability of humanoid robots, along with the necessity to optimize a composite reward structure to ensure task performance (\textit{e.g.}, tracking accuracy) and reliability in real-world deployment (\textit{e.g.}, energy efficiency), significantly increase the sample complexity. 

In response to this challenge, recent work has focused on simulation acceleration \cite{makoviychuk2021isaac,Mayank2023Orbit,mujoco_mjx}, data augmentation \cite{schwarke2023curiosity,mittal2024symmetry,schwarke2025rslrl}, and state representation learning (SRL) \cite{long2024learning,sun2025learning,zhang2025track}. Among these strategies, SRL offers a promising solution by transforming high-dimensional sensory inputs into compact and informative latent representations. These representations preserve task-relevant dynamics while filtering out noise and redundancy, which can significantly improve sample efficiency, generalization, and overall performance in RL \cite{laskin2020curl,schwarzer2021dataefficient,laskin2022unsupervised,echchahed2025a} tasks. Pioneering work like \cite{long2024learning,sun2025learning} demonstrates the potential of SRL in humanoid robot learning, where reconstruction-based representations (\textit{e.g.}, elevation-based internal maps \cite{sun2025learning}) and contrastive learning-based abstractions (\textit{e.g.}, perceptive internal models from height maps \cite{long2024learning}) both serve to refine state embeddings and improve RL performance in complex environments. However, integrating SRL techniques into humanoid WBC remains underexplored, especially the seamless combination of SRL with RL in a unified, end-to-end framework that enhances both learning efficiency and real-world deployment reliability.

In this paper, we investigate SRL-empowered humanoid robot learning and propose a simple yet powerful approach entitled \textbf{PvP}: \textbf{P}roprioceptive-\textbf{P}rivileged contrastive learning. Our contributions are threefold:

\begin{itemize}
    \item We propose to conduct contrastive learning between proprioceptive and privileged states of the humanoid robot to enhance the proprioceptive representations for policy learning. PvP leverages the intrinsic complementarity between the two state modalities without relying on hand-crafted data augmentations, producing stable improvements across a broad range of tasks.
    
    \item We introduce \textbf{SRL4Humanoid}, to the best of our knowledge, the first unified and modular open-source framework that provides high-quality implementations of representative SRL methods for humanoid robot learning, enabling reproducible research and facilitating future progress in the community. Equipped with SRL4Humanoid, we conduct a systematic study of how different SRL methods and configurations affect the efficiency and performance of humanoid WBC learning.
    
    \item We validate our approach on the \textbf{LimX Oli} humanoid robot through two representative tasks: \textbf{velocity tracking} and \textbf{motion imitation}. Extensive experiment results demonstrate that PvP significantly enhances sample efficiency and policy performance in complex WBC scenarios, outperforming existing SRL baselines.
\end{itemize}

\section{Related Work}

\subsection{SRL for Reinforcement Learning}
SRL has been widely applied in RL to enhance learning efficiency, with predominant approaches broadly categorized into three types: reconstruction-based \cite{kingma2014auto,higgins2017beta,yu2022mask,dunion2023temporal,dunion2023conditional}, dynamics modeling \cite{pathak2017curiosity,islam2023principled,lamb2025guaranteed,schwarzer2021dataefficient,mcinroe2025multi}, and contrastive learning techniques \cite{stooke2021decoupling,yu2021playvirtual,laskin2022unsupervised,laskin2022cic,zheng2024contrastive}. For instance, dynamics modeling methods encourage representations that capture environment dynamics through predictive modeling. Forward models predict future states from current state–action pairs, while inverse models infer actions from transitions \cite{pathak2017curiosity,islam2023principled,lamb2025guaranteed}, thereby providing rich controllable features. In contrast, contrastive learning methods structure latent space by enforcing similarity between positive pairs (\textit{e.g.}, temporally adjacent states) while separating negative pairs \cite{oord2018representation}. This yields invariant and temporally smooth representations that improve RL efficiency and generalization. For instance, CURL \cite{laskin2022unsupervised} leverages augmented image pairs to enforce invariances, while ATC \cite{stooke2021decoupling} aligns embeddings of temporally close observations. Extensions such as CDPC \cite{zheng2024contrastive} refine contrastive predictive coding with temporal-difference objectives to stabilize training in stochastic environments.

\subsection{SRL for Humanoid Robot Learning}\label{sec:related srl}
SRL holds great promise for enhancing humanoid robot learning by enabling efficient encoding and processing of complex sensory information, with pioneering work exploring its application in improving robot adaptability and performance in dynamic environments \cite{long2024learning,sun2025learning,sun2025learning_world,zhang2025track,zheng2025transformer}. For example, \cite{zhang2025track} proposes an Any2Track framework that leverages SRL to enhance motion tracking by integrating a history-informed adaptation module. This module utilizes dynamics-aware world model prediction to extract informative dynamics features, enabling the robot to adapt to various disturbances, including terrain, external forces, and changes in physical properties. Similarly, \cite{sun2025learning_world} proposed a world model reconstruction framework that uses sensor denoising and world state estimation to improve locomotion in unpredictable environments. These approaches highlight the potential of SRL to enhance the performance of humanoid robots, yet further exploration is needed to optimize and generalize these methods across diverse control tasks and environments.

\section{Background}
\subsection{Humanoid Whole-Body Control}
Whole-body control (WBC) serves as the foundation for enabling humanoid robots to execute diverse and complex tasks. Formally, given a set of continuous commands $\mathcal{C}$ and observations $\mathcal{O}$, the objective is to design a control function that maps $(\mathcal{O}, \mathcal{C})$ to appropriate control signals. Equipped with learning-based methods, it is feasible to directly learn a parameterized policy $\pi_{\bm\theta}: \mathcal{O} \times \mathcal{C} \rightarrow \mathcal{A}$ that outputs joint actions \cite{siekmann2021sim}. In practice, actions are often defined as offsets to nominal joint positions for the upper body, lower body, and hands. The final joint reference is obtained by adding these offsets to nominal targets, which are then tracked using a proportional-derivative (PD) controller with fixed gains.

\subsection{Reinforcement Learning}
RL offers a data-driven framework for optimizing humanoid WBC policies through interaction with the environment. Formally, humanoid WBC can be cast as an infinite-horizon partially observable Markov decision process (POMDP) \cite{spaan2012partially} defined by the tuple $\mathcal{M}=(\mathcal{S}, \mathcal{O}, \mathcal{A}, P, \Omega, R, \gamma)$. Here, $\mathcal{S}$ denotes the full state space of the humanoid and its environment, $\mathcal{O}$ is the observation space, $\mathcal{A}$ is the action space, $P(\bm{s}'|\bm{s},\bm{a})$ is the transition probability function, $\Omega(\bm{o},\bm{s},\bm{a})$ is the observation function, $R(\bm{s},\bm{a},\bm{s}')$ is the reward function, and $\gamma\in[0,1]$ is a discount factor. The objective of RL is to learn an optimal policy $\pi_{\bm\theta}$ that maximizes the expected discounted return:

\begin{equation}
    J_\pi(\bm{\theta}) = \mathbb{E}_\pi \Big[ \sum_{t=0}^{\infty} \gamma^t R(\bm{s}_t, \bm{a}_t, \bm{s}_{t+1}) \Big].
\end{equation}

\textbf{Proprioceptive State Space}. Denote by $\bm{o}_t \in \mathbb{R}^n$ the proprioceptive state of the robot at time step $t$, which consists of signals directly measurable on hardware, typically joint positions $\bm{q}_t$, joint velocities $\dot{\bm{q}}_t$, base angular velocity $\bm{\omega}_t$, and the gravity (orientation) $\bm{g}_t$ estimate in the base frame.

\textbf{Privileged State Space}. At time step $t$, $\bm{s}_t \in \mathbb{R}^m$ denotes the full simulator state used only during training (\textit{e.g.}, by the critic/teacher) but unavailable or unreliable on the real robot. Typical components include root pose and velocity, per-link poses and velocities, contact indicators, and environment/terrain features. Notably, we let the privileged state $\bm{s}$ and proprioceptive state $\bm{o}$ satisfy $\bm{o}\subset\bm{s}$.

\textbf{Action Space.} The action $\bm{a}_t \in \mathbb{R}^k$ specifies angular deviations for $k$ actuated joints relative to their nominal positions. The final target joint positions are computed by adding $\bm{a}_t$ to the nominal configuration, which is then tracked by a PD controller.

\section{Methodology}\label{sec:srl4hum}
In this section, we introduce \textbf{PvP}: \textbf{P}roprioceptive-\textbf{P}rivileged contrastive learning, a general, simple, yet powerful framework to accelerate the learning of humanoid WBC tasks. We also present \textbf{SRL4Humanoid}, a unified, modular, and plug-and-play toolkit that provides high-quality implementations of representative SRL methods for humanoid robot learning.

\subsection{PvP Implementation}

\begin{figure}[h!]
    \centering
    \includegraphics[width=\linewidth]{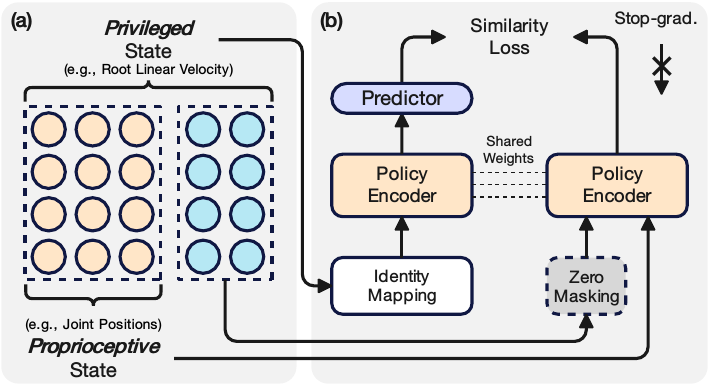}
    \caption{An overview of the PvP approach. (a) The components of the privileged state and the proprioceptive state. (b) PvP conducts contrastive learning based on the intrinsic complementarity between the two state modalities.}
    \label{fig:pvp}
\end{figure}


Recent approaches in humanoid robot learning have introduced SRL as an auxiliary learning objective to enhance policy learning. However, most of them follow a reconstruction-based approach \cite{wang2024cts,sun2025learning_world} (\textit{e.g.}, predicting privileged information such as root linear velocity from proprioceptive states), yet they often struggle with suboptimal representation quality and poor generalization. This is because these methods focus on preserving the complete state, including irrelevant details, rather than learning task-relevant features. While methods like PIM \cite{long2024learning} employ contrastive learning to obtain more robust representations, they only rely on a single state modality without incorporating privileged state information. This limitation limits their ability to capture the full spectrum of task-relevant dynamics, leading to less-informed representations.

With this in mind, PvP utilizes both proprioceptive and privileged states to perform contrastive learning, combining complementary information from different sensory modalities. As shown in Figure~\ref{fig:pvp}, the privileged state $\bm{s}$ contains both proprioceptive observations (\textit{e.g.}, joint positions and angular velocities) and privileged information (\textit{e.g.}, root linear velocity), which can be considered as the pseudo augmentation of the proprioceptive state $\bm{o}$. Meanwhile, we apply the zero-masking to the privileged information part in $\bm{s}$ and keep the proprioceptive observations solely:
\begin{equation}
    \tilde{\bm{s}}_{t}=\text{ZeroMasking}(\bm{s}_t).
\end{equation}
Then, the derived data pair $(\bm{s},\tilde{\bm s})$ is used to train the policy encoder, following the SimSiam algorithm \cite{Chen2021Simsiam}. Formally, denote by $f_{\bm\theta}$ the policy encoder, $h_{\bm\psi}$ the predictor, we have
\begin{equation}
\begin{aligned}
    \bm{z}=f_{\bm\theta}(\bm{s})&,\quad \tilde{\bm{z}}=f_{\bm\theta}(\tilde{\bm{s}})\\
    \bm{p}=h_{\bm\psi}\left(\bm{z}\right)&,\quad \tilde{\bm{p}}=h_{\bm\psi}\left(\tilde{\bm{z}}\right)
\end{aligned}
\end{equation}

Finally, the PvP loss is defined as
\begin{equation}
    L_{\rm PvP}=D_{\rm ncs}\left(\bm{p},\text{sg}(\tilde{\bm{z}})\right)+D_{\rm ncs}\left(\tilde{\bm{p}},\text{sg}(\bm{z})\right),
\end{equation}

where $D_{\rm ncs}(\bm{p},\bm{z})=-\frac{\bm p}{\Vert\bm{p}\Vert_{2}}\cdot \frac{\bm z}{\Vert\bm{z}\Vert_{2}} $ is the negative cosine similarity loss between the pair and $\text{sg}(\cdot)$ is the stop-gradient operation.

We would like to highlight the advantages of PvP. First, PvP leverages both proprioceptive and privileged states for contrastive learning, effectively reducing the complexity of SRL while enhancing the learned representations by incorporating richer and more comprehensive information. This approach also offers an alternative method for the policy to access privileged information, thereby enabling the agent to gain a better understanding of its environment. Additionally, PvP leverages the intrinsic complementarity between the two state modalities without relying on hand-crafted data augmentations. As a result, PvP is highly versatile and can be applied to a wide range of tasks.

\subsection{The SRL4Humanoid Framework}\label{sec:toolkit}

\begin{figure}[h!]
    \centering
    \includegraphics[width=\linewidth]{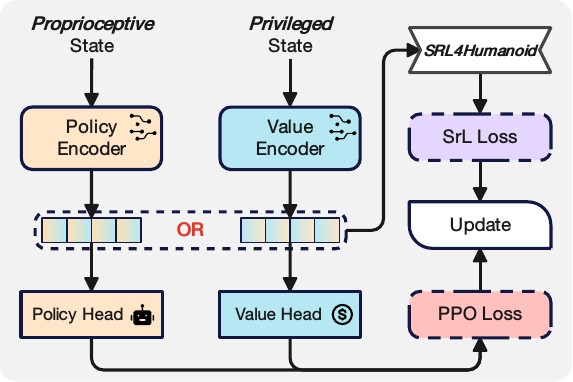}
    \caption{The architecture of the SRL4Humanoid framework, in which the SRL and RL processes are fully decoupled.}
    \label{fig:srl4hum}
\end{figure}

To facilitate our experiments and subsequent research, we develop SRL4Humanoid, a unified, highly modularized framework that provides high-quality and reliable implementations of representative SRL techniques for humanoid robot learning. Figure~\ref{fig:srl4hum} illustrates the architecture of the SRL4Humanoid framework. Following the practice of a series of humanoid robot research \cite{cheng2024expressive,long2024learning,sun2025learning,xue2025unified}, we select the proximal policy optimization (PPO) \cite{schulman2017proximal} as the backbone RL algorithm. Notably, the policy network accepts the proprioceptive state of the robot to generate actions, while the value network accepts the privileged state of the environment to perform value estimation. The SRL and RL processes are designed to be fully decoupled, and depending on the training configuration, the SRL objective can be applied to either the policy encoder or the value encoder.

To ensure both diversity and representativeness, we currently implement three widely studied SRL algorithms, each exemplifying a distinct methodological paradigm introduced in Section~\ref{sec:related srl}. These categories capture the main approaches explored in prior SRL research, allowing us to analyze their relative strengths when applied to humanoid WBC tasks. Finally, the joint optimization objective of the RL and the SRL is defined as
\begin{equation}\label{eq:total loss}
    L_{\rm Total}=L_{\rm RL}+\lambda\cdot L_{\rm SRL},
\end{equation}
where $\lambda$ is a weighting coefficient. 

\begin{algorithm}[t!]
\SetAlgoLined
\caption{Workflow of the SRL4Humanoid}
\label{algo:srl4hum}
Initialize the policy $\pi_{\bm \theta}$ and value network $V_{\bm \phi}$;

Initialize the SRL module $S_{\bm\psi}$;

Set all the hyperparameters, such as the maximum number of episodes $E$, and the number of update epochs $K$, etc.

\For{episode = 1 \KwTo $E$}{
Sample rollouts using the policy network $\pi_{\bm \theta}$;

Perform the generalized advantage estimation (GAE) to get the estimated task returns;

\For{epoch = $1$ \KwTo $K$}{
Sample a mini-batch $\mathcal{B}$ from the rollouts data;

Use $\mathcal{B}$ to compute the policy and value loss;

Use $\mathcal{B}$ to compute the SRL loss;

Compute the total loss following Eq.~(\ref{eq:total loss interval});

Update the policy network, value network, and the SRL module;
}
}
Output the optimized policy $\pi_{\bm \theta}$.
\end{algorithm}

By default, the updates of the two loss items are synchronized. This means that the training of the SRL module shares the data batches with RL and follows the update frequency of RL to isolate the effects of SRL. However, in practical experiments, we found that consistently applying the SRL loss does not always have a positive influence on policy learning and can sometimes degrade learning efficiency. In massively parallel RL, a large amount of repetitive and low-quality data is produced during the early stages of training. This is likely to cause the SRL to prematurely fall into local optima and fail to continuously influence policy learning. To address this issue, we employ an interval update mechanism:
\begin{equation}\label{eq:total loss interval}
    L_{\rm Total}=L_{\rm RL}+\mathbbm{1}(T) \cdot\lambda\cdot L_{\rm SRL},
\end{equation}

where $\mathbbm{1}(T)$ is an indicator function that equals $1$ every $T$ time steps; otherwise, it equals $0$. This phenomenon will be detailed in the experiment section. Finally, we summarize the workflow of SRL4Humanoid in Algorithm~\ref{algo:srl4hum}.

\section{Experiments}

\begin{figure*}[ht!]
    \centering
    \includegraphics[width=\linewidth]{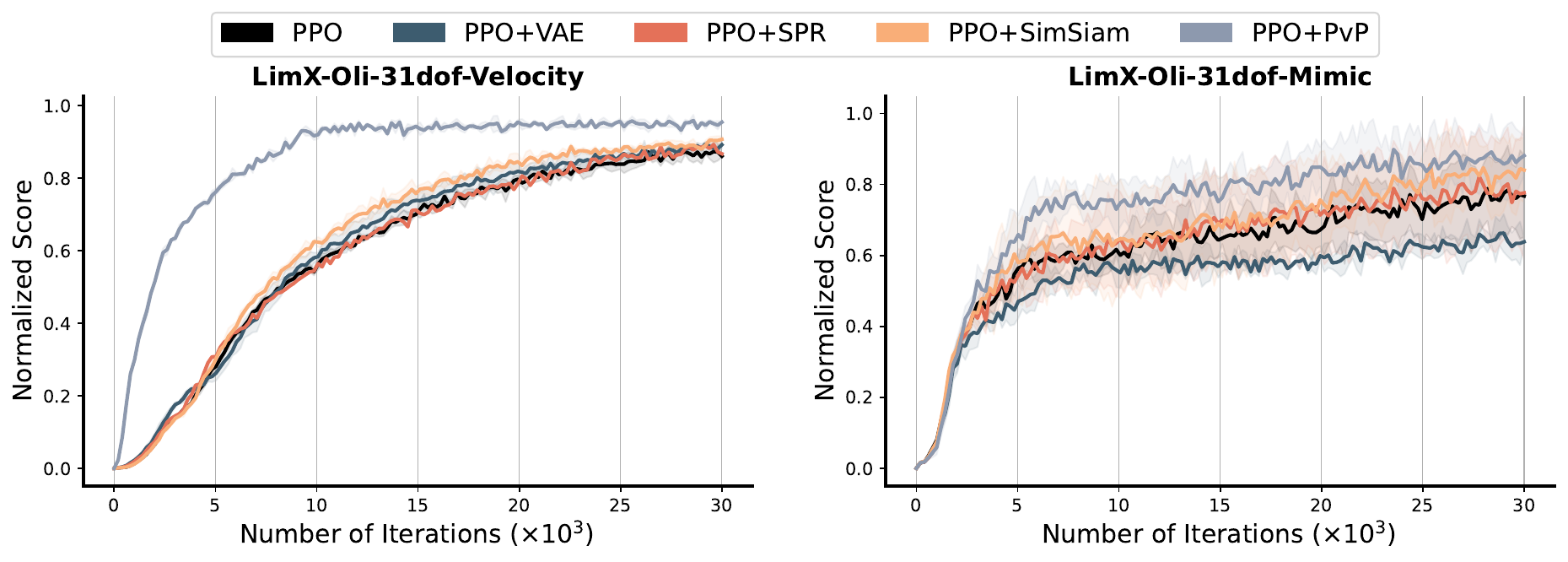}
    \vspace{-20pt}
    \caption{Training progress comparison between the vanilla PPO agent and its combination with four SRL methods on the two humanoid WBC tasks. The solid line and shaded region denote the mean and standard deviation, respectively.}
    \label{fig:q1_overall_score}
    \vspace{-10pt}
\end{figure*}

In this section, we design the experiments to achieve the two main objectives: (i) evaluate the performance of the proposed PvP algorithm and the SRL4Humanoid framework on challenging humanoid robot tasks, and (ii) conduct a systematic analysis of the application of SRL for humanoid WBC. Specifically, we formulate the following research questions:
\begin{itemize}
    \item \textbf{Q1}: Can the proposed PvP algorithm outperform the baseline methods? (See Figure~\ref{fig:q1_overall_score}, \ref{fig:q1_action_smoothness}, \ref{fig:q1_tracking_score})
    \item \textbf{Q2}: How does the proportion of training time affect SRL's performance? (See Figure~\ref{fig:q2_time_prop})
    \item \textbf{Q3}: How does the proportion of training data affect SRL's performance? (See Figure~\ref{fig:q2_data_prop})
    \item \textbf{Q4}: Which encoder (policy or value) benefits more from applying SRL loss? (See Figure~\ref{fig:q4_encoder})
    \item \textbf{Q5}: How computation-efficient are these SRL methods? (See Section~\ref{sec:fps})
    \item \textbf{Q6}: How do the SRL-enhanced methods behave in real-world deployment? (See Figure~\ref{fig:big_preface})
\end{itemize}

\begin{figure}[ht!]
    \centering
    \includegraphics[width=\linewidth]{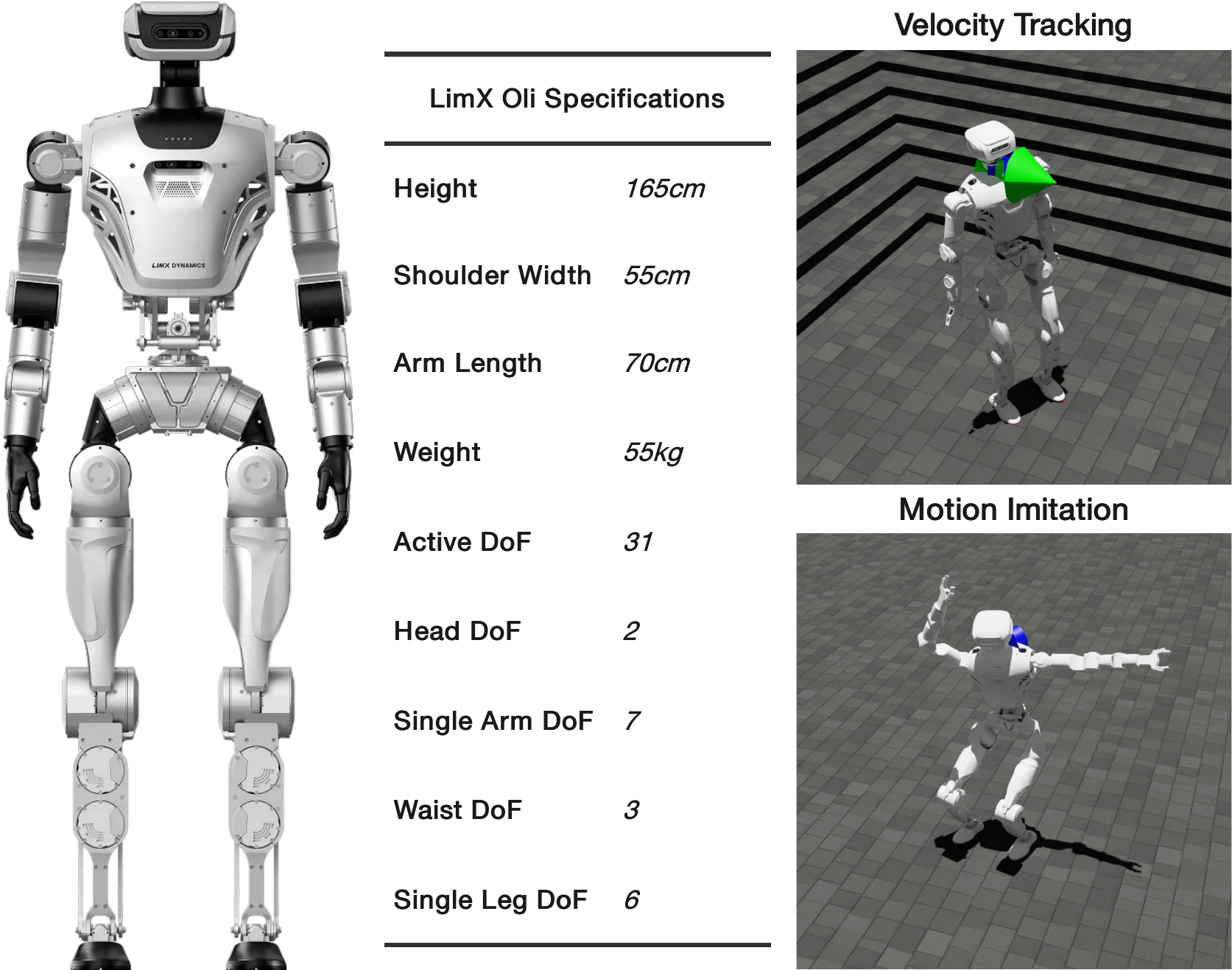}
    \caption{The specifications of the LimX Oli humanoid robot used in the experiments, and the screenshots of the two designed tasks.}
    \label{fig:screenshots_specs}
    \vspace{-15pt}
\end{figure}

\subsection{Experimental Setup}

\subsubsection{Benchmark Design} As shown in Figure~\ref{fig:screenshots_specs}, we utilize the LimX Oli as the test platform, which is a full-size humanoid robot with 31 degrees of freedom (DoF). Based on this platform, we design two tasks: \textit{LimX-Oli-31dof-Velocity} and \textit{LimX-Oli-31dof-Mimic}. Specifically, \textit{LimX-Oli-31dof-Velocity} requires the robot to track a velocity command on flat terrain, which is resampled every $10$ seconds. The range of linear velocity on the $x$-axis is $(-0.5, 1.0)$ m/s, $(-0.3, 0.3)$ m/s on the $y$-axis, and the angular velocity on the $z$-axis is $(-1.0, 1.0)$ rad/s. For \textit{LimX-Oli-31dof-Mimic}, it requires the robot to imitate different pre-recorded human animations. We use a set of $20$ pre-recorded human motions (Figure~\ref{fig:mocap}), each with a maximum length of 43 seconds and 4,300 frames. The two tasks encompass the primary categories of evaluation benchmarks in current humanoid robot research \cite{xue2025unified}. More details about the task configuration can be found in Appendix~\ref{appendix:benchmark}.

\subsubsection{Algorithmic Baselines}
As introduced in Section~\ref{sec:toolkit}, we use PPO as the backbone RL algorithm and combine it with different SRL approaches. For SRL approaches, we implement SimSiam \cite{Chen2021Simsiam}, SPR \cite{schwarzer2021dataefficient}, and VAE \cite{kingma2014auto}. The details of the selected algorithmic baselines can be found in Appendix~\ref{appendix:baselines}. By default, these SRL loss terms are applied to the policy encoder. We also conduct a hyperparameter search for these methods to determine the initial hyperparameters, such as data augmentation operations (\textit{e.g.}, Gaussian noise and random masking) and loss coefficients. To make a fair comparison, all the methods share the same network architectures and training steps. More details of the experimental setup can be found in Appendix~\ref{appendix:setup}.
\subsubsection{Evaluation Metrics}
The evaluation metrics are fourfold. First, we compare the overall task performance of all the methods, which are computed as the weighted summation of all the sub-reward functions. Second, we compare the key performance indicators (KPIs) for each task, including velocity tracking accuracy in \textit{LimX-Oli-31dof-Velocity} and position alignment accuracy in \textit{LimX-Oli-31dof-Mimic}. Furthermore, we compare the training efficiency of all the methods, \textit{i.e.}, if the method can achieve a higher convergence speed. Finally, we also consider the effectiveness of learned policies in real-world deployment. For example, the agent must effectively optimize the action smoothness reward term to prevent the robot from moving violently during real-time control.

\subsection{Result Analysis}

The following results analysis is performed based on the pre-defined research questions. More ablation studies can be found in Appendix~\ref{sec:ablation}.

\subsubsection{Task Performance Comparison}

\begin{figure}[h!]
    \centering
        \includegraphics[width=\linewidth]{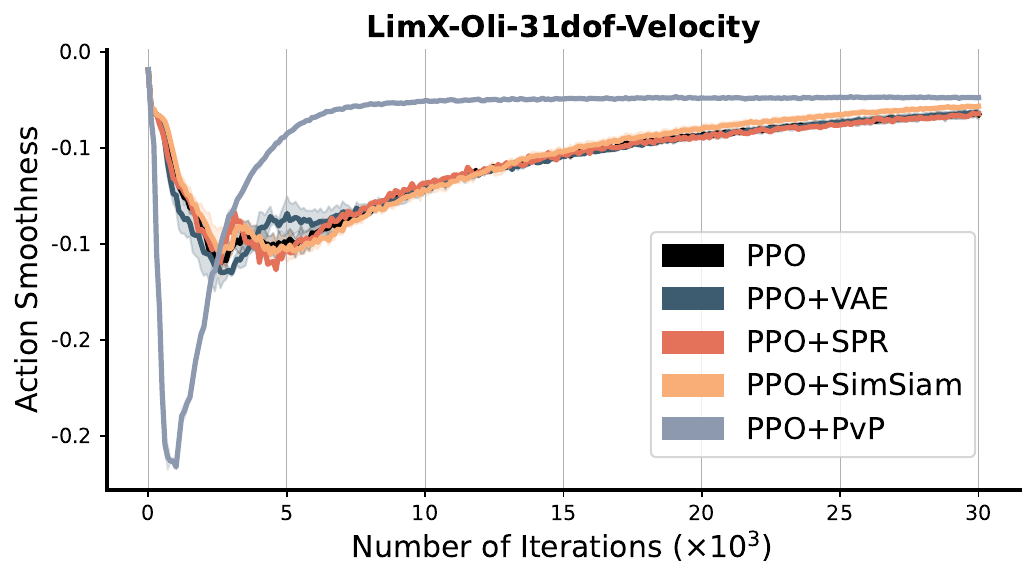}
        \vspace{-20pt}
        \caption{The comparison of action smoothness optimization between the vanilla PPO agent and its combinations with the four SRL methods. The solid line and shaded region denote the mean and standard deviation, respectively.}
        \label{fig:q1_action_smoothness}
        \vspace{-15pt}
\end{figure}

We first analyze the overall effect of integrating SRL into the WBC learning process, \textit{i.e.}, the accumulation of all reward terms. Figure~\ref{fig:q1_overall_score} illustrates the overall reward comparison between the vanilla PPO agent and its combination with the four SRL methods on the two humanoid WBC tasks. For the velocity tracking task, our PvP method significantly accelerates the learning process, while other SRL methods produce marginal improvements in learning speed. This demonstrates the advantage of leveraging privileged information to enhance SRL, allowing the agent to extract more informative features from noisy and redundant sensory inputs, thereby accelerating the learning process. In contrast, for the motion imitation task, two of the SRL methods outperform the vanilla PPO agent, with PvP achieving the highest performance. However, the VAE method exhibits performance degradation, indicating that simply reconstructing sensory data is insufficient to enhance the robot's learning efficiency. These results demonstrate that learning high-quality features can improve both learning efficiency and the final performance in humanoid WBC tasks.

\begin{figure}[h!]
    \centering
        \includegraphics[width=\linewidth]{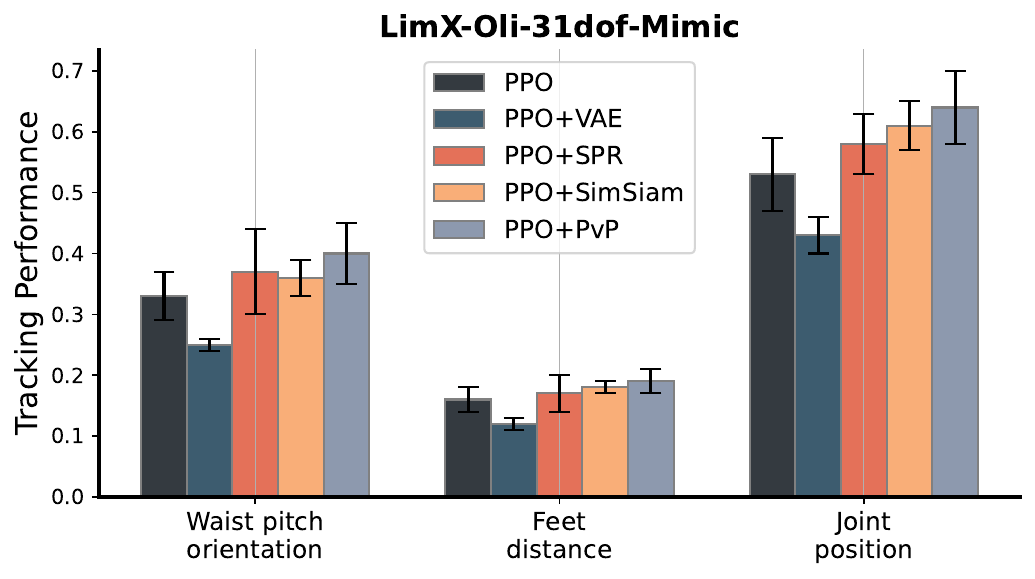}
        \vspace{-20pt}
        \caption{The tracking performance comparison between the PPO agent and its combinations with the four SRL methods. Our PvP achieves the highest performance across the three key tracking metrics.}
        \label{fig:q1_tracking_score}
        \vspace{-15pt}
\end{figure}

Next, we evaluate the performance of these methods based on specific reward terms. For the velocity tracking task, we compare the optimization of the action smoothness term, which is defined as the difference between the robot's actions across three consecutive frames. This term helps prevent abrupt robot movements, ensuring smoother, more controlled motions during real-world deployment. As shown in Figure~\ref{fig:q1_action_smoothness}, PvP significantly accelerates the convergence of this penalty term. This indicates that our PvP method can not only accelerate the policy learning in simulation but also guarantee the reliability of real-world deployment. In contrast, the motion imitation task places higher demands on control precision. Figure~\ref{fig:q1_tracking_score} illustrates the comparison of PvP and other baseline algorithms across three key metrics. It is clear that the PvP approach not only improves overall performance but also optimizes critical KPIs, providing reliable increments across diverse tasks.

\subsubsection{Impact of Training Time Proportion}

\begin{figure}[h!]
    \vspace{-10pt}
    \centering
    \begin{subfigure}{\linewidth}
        \centering
        \includegraphics[width=\textwidth]{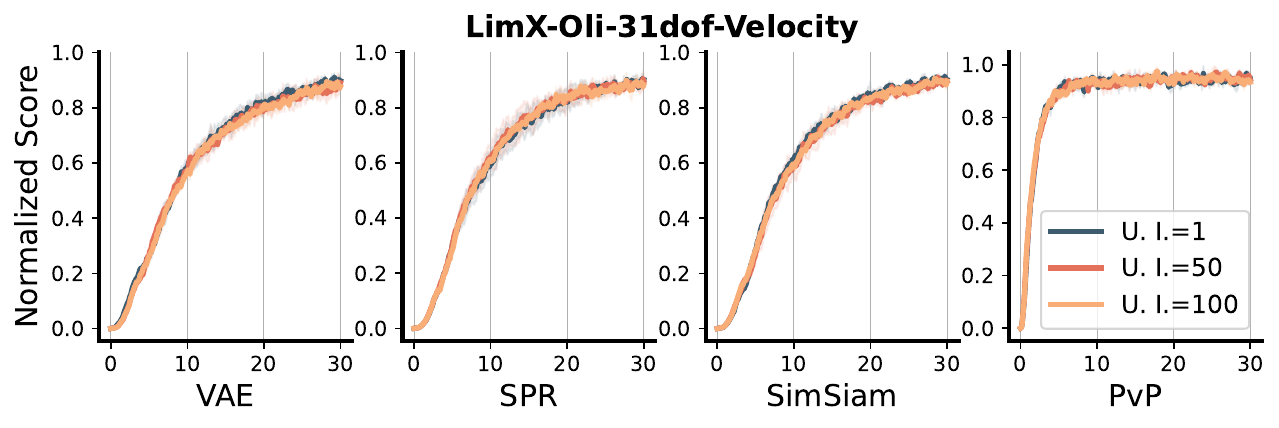}
    \end{subfigure}
    \vfill
    \begin{subfigure}{\linewidth}
        \centering
        \includegraphics[width=\textwidth]{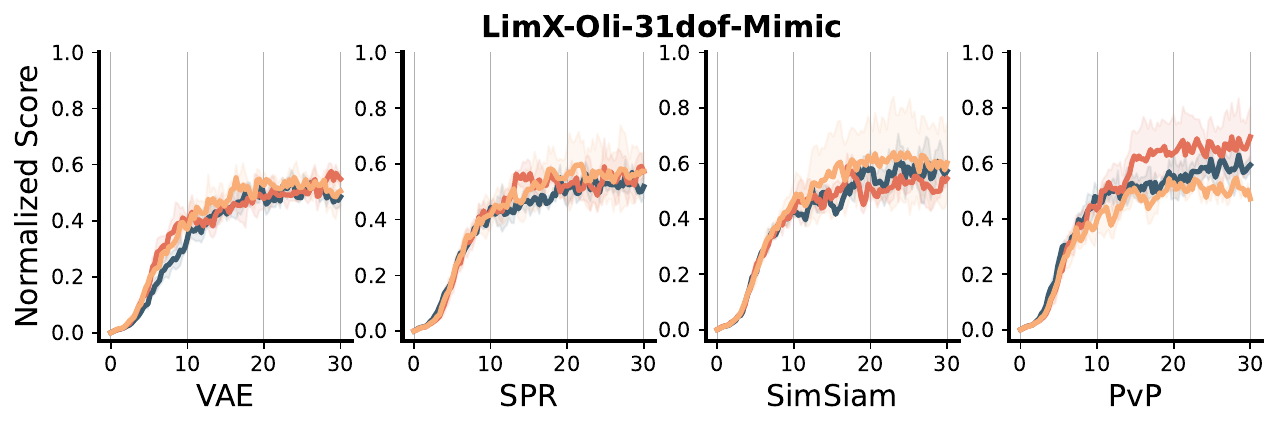}
    \end{subfigure}
    \vspace{-20pt}
    \caption{Training progress comparison of the four SRL methods with different training time proportions on the two humanoid WBC tasks. The solid line and shaded region denote the mean and standard deviation, respectively.}
    \vspace{-10pt}
    \label{fig:q2_time_prop}
\end{figure}

All the SRL methods are designed to be updated synchronously with the RL part as an auxiliary task. In the early stages of training, SRL helps the agent quickly understand its environment by providing informative state representations. However, as training progresses and the agent becomes more adept at the task, the need for continuous SRL updates diminishes. Additionally, during the initial stages, large amounts of homogeneous and low-quality data in the early stage are likely to cause the SRL module to prematurely fall into local optima. Thus, it is worthwhile to investigate how the proportion of training time impacts SRL's performance. 

We evaluate different update intervals (1, 50, and 100) for each SRL method. As shown in Figure~\ref{fig:q2_time_prop}, adjusting the update intervals has minimal effect on the velocity tracking task, but a clear impact on the motion imitation task. Specifically, an update interval of 50 is generally optimal for all the SRL methods. These results indicate that carefully selecting the update interval can improve the performance of SRL, prevent premature convergence to local optima, and boost overall training efficiency by reducing computational overhead.

\subsubsection{Impact of Training Data Proportion}

\begin{figure}[h!]
    \vspace{-10pt}
    \centering
    \begin{subfigure}{\linewidth}
        \centering
        \includegraphics[width=\textwidth]{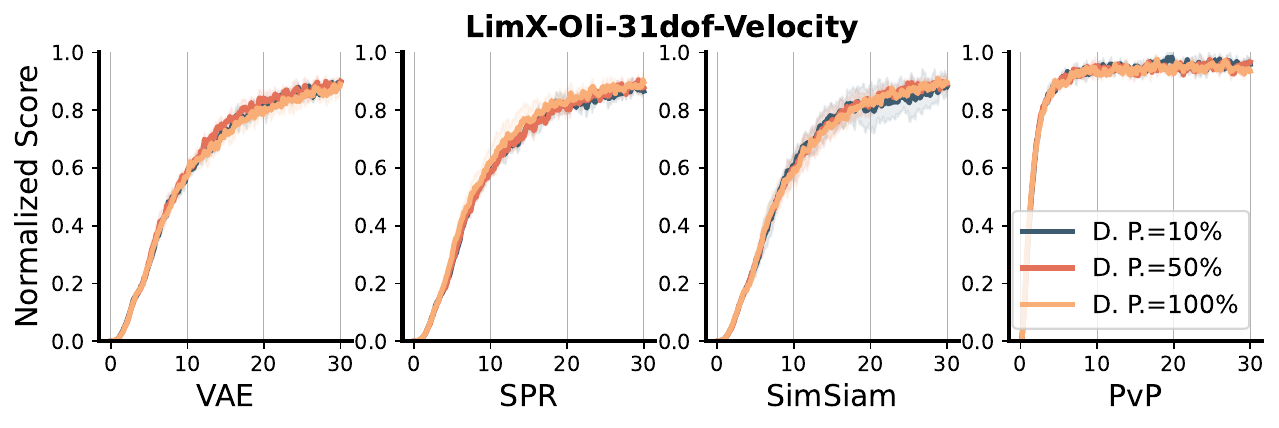}
    \end{subfigure}
    \vfill
    \begin{subfigure}{\linewidth}
        \centering
        \includegraphics[width=\textwidth]{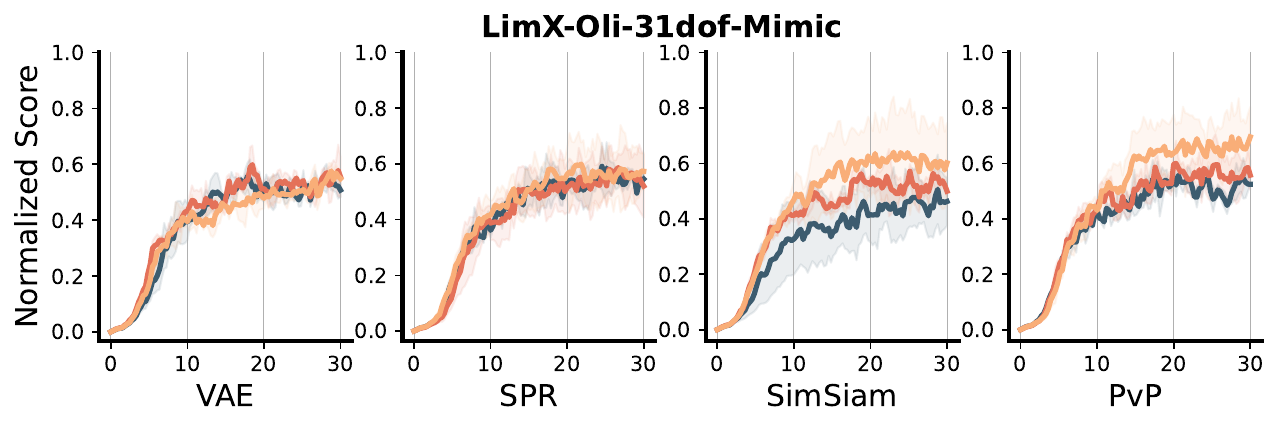}
    \end{subfigure}
    \vspace{-20pt}
    \caption{Training progress comparison of the four SRL methods with different training data proportions on the two humanoid WBC tasks. The solid line and shaded region denote the mean and standard deviation, respectively.}
    \vspace{-10pt}
    \label{fig:q2_data_prop}
\end{figure}

Likewise, we examine the impact of the proportion of training data on SRL's performance, inspired by \cite{yuan2025rlexplore}, which investigates the impact of the training data proportion on intrinsic rewards. We train each SRL method using 10\%, 50\%, and 100\% of the rollouts data sampled by the agent in each episode, with data resampled through a random masking operation. Figure~\ref{fig:q2_data_prop} illustrates the training progress comparison of the four SRL methods across different training data proportions. For the velocity-tracking task, using different proportions yields nearly identical training curves. In contrast, increasing the proportion generally enhances performance, particularly for the SimSiam and PvP methods. These results demonstrate that allocating an appropriate proportion of training data accelerates learning and improves performance, especially in the motion imitation task.

\begin{figure}[ht!]
    \centering
    \begin{subfigure}{\linewidth}
        \centering
        \includegraphics[width=\textwidth]{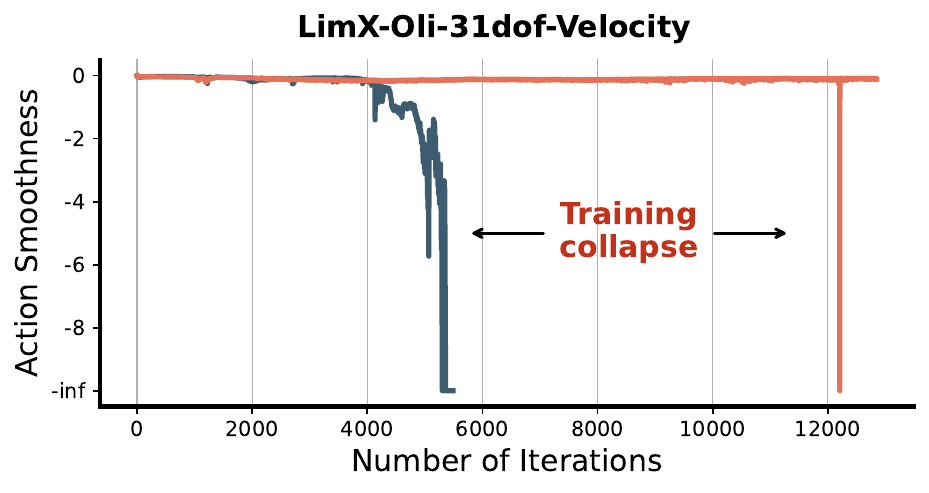}
    \end{subfigure}
    \vfill
    \begin{subfigure}{\linewidth}
        \centering
        \includegraphics[width=\textwidth]{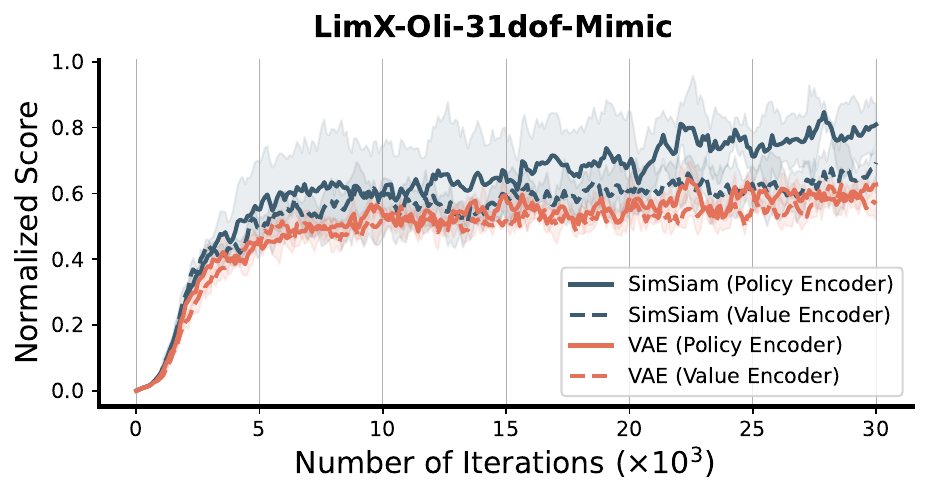}
    \end{subfigure}
    \vspace{-20pt}
    \caption{Learning curves of applying the SRL to the value encoder. The solid line and shaded region denote the mean and standard deviation, respectively.}
    \label{fig:q4_encoder}
    \vspace{-10pt}
\end{figure}

\subsubsection{Impact of SRL on Value Encoder}
Previous experiments have focused on applying SRL to the policy encoder to improve the understanding of the proprioceptive state. Correspondingly, it is worth exploring the impact of SRL on the value encoder. We implement this functionality in SRL4Humanoid and conduct ablation experiments using combinations of the PPO agent and two SRL methods (SPR requires state-action pairs for training). As shown in Figure~\ref{fig:q4_encoder}, applying SRL loss to the value encoder leads to slower convergence compared to applying SRL loss to the policy encoder. Specifically, in the velocity tracking task, we observe a collapse in training when SRL is applied to the value encoder, as indicated by the sharp drop in action smoothness before recovery. These findings suggest that applying SRL to the policy encoder results in a more stable learning process and improved performance.

\subsubsection{Computational Efficiency Comparison}\label{sec:fps}


Furthermore, we investigate the computational efficiency of these SRL methods. Equipped with IsaacLab \cite{nvidia2025isaaclab}, all experiments can be run on a single GPU, and a RTX4090 with 24GB memory is used for all our experiments. Our implementation enables the SRL module to run entirely on the GPU, without affecting overall training efficiency. We have attached the complete Weights \& Biases (Wandb) logs to the compute reporting form (CRF). These results demonstrate that SRL4Humanoid can effectively accelerate humanoid WBC tasks with minimal computational resource cost.

\subsubsection{Real-world Evaluation}
\begin{figure}[h!]
    \vspace{-5pt}
    \centering
    \begin{subfigure}{\linewidth}
        \centering
        \includegraphics[width=\textwidth]{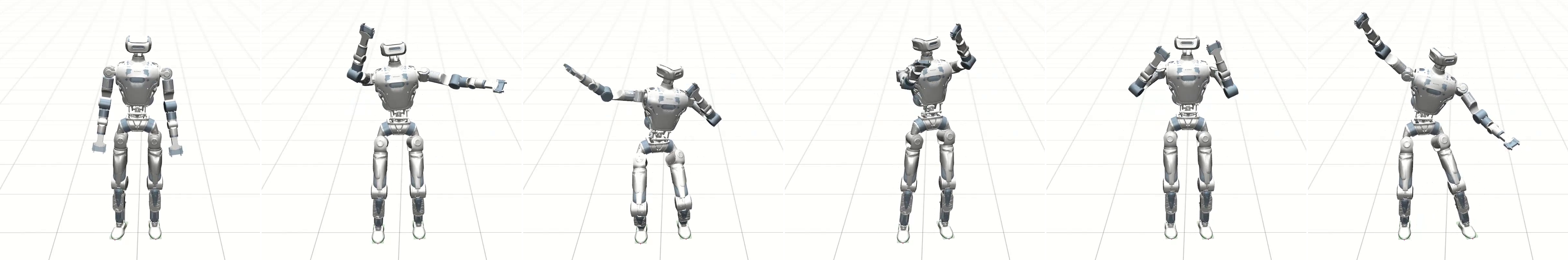}
    \end{subfigure}
    \vfill
    \begin{subfigure}{\linewidth}
        \centering
        \includegraphics[width=\textwidth]{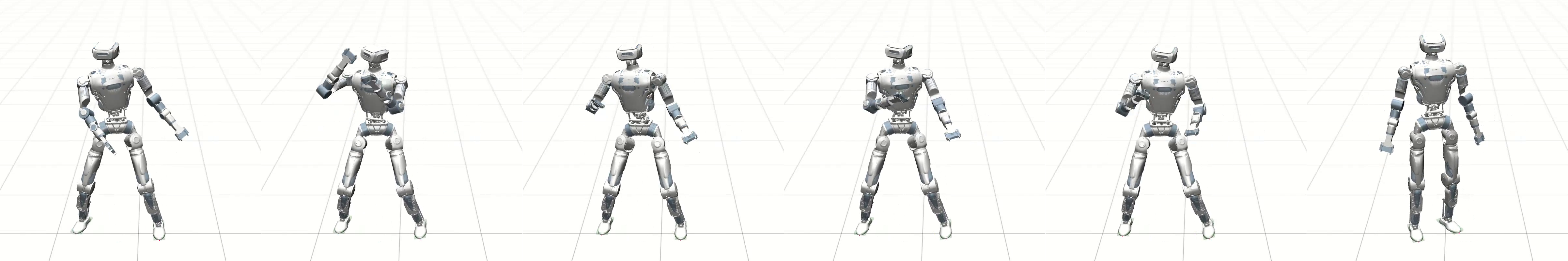}
    \end{subfigure}
    \vfill
    \begin{subfigure}{\linewidth}
        \centering
        \includegraphics[width=\textwidth]{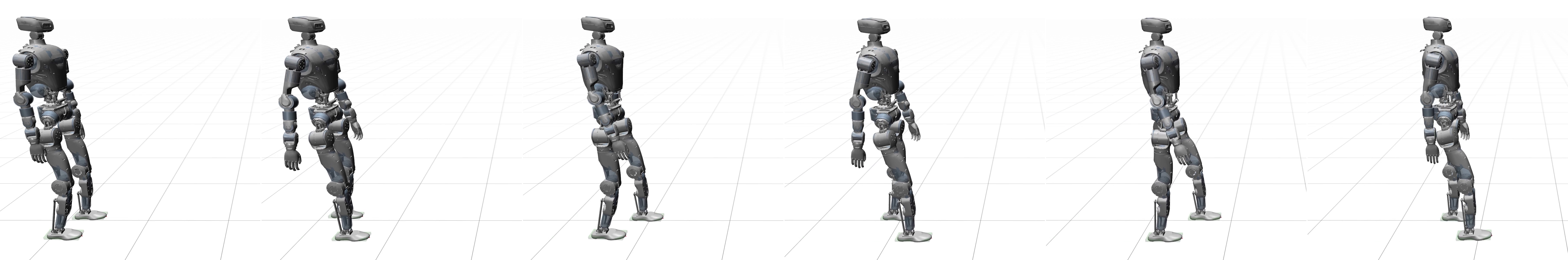}
    \end{subfigure}
    \vfill
    \begin{subfigure}{\linewidth}
        \centering
        \includegraphics[width=\textwidth]{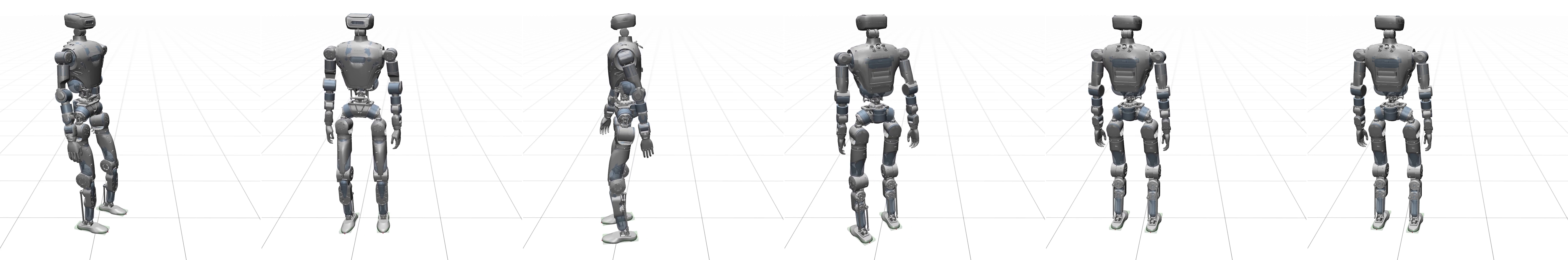}
    \end{subfigure}
    \vspace{-10pt}
    \caption{Sim2Sim evaluation on the MuJoCo simulator. The first two rows demonstrate motion imitation ability, and the last two rows show velocity tracking ability.}
    \label{fig:mujoco_eval}
\end{figure}

To assess the effectiveness of our approach in real-world scenarios, we first conduct a thorough Sim2Sim evaluation on the MuJoCo platform \cite{todorov2012mujoco}, which provides simulation precision closer to real-world conditions than IsaacLab \cite{Mayank2023Orbit}. Figure~\ref{fig:mujoco_eval} provides a clear demonstration of the robot’s ability to execute complex tasks with the learned policy. Following the simulation evaluation, we proceed with real-robot testing on the LimX Oli humanoid robot, as shown in Figure~\ref{fig:big_preface}. More demonstrations can be found in the Supporting Videos in the supplementary materials.

\vspace{-10pt}

\section{Conclusion}
In this paper, we propose PvP, a proprioceptive-privileged contrastive learning framework that enhances sample efficiency and performance in humanoid WBC tasks. By leveraging the intrinsic complementarity of the two state modalities, PvP enhances proprioceptive representations for policy learning without relying on hand-crafted data augmentations. We also introduce the SRL4Humanoid framework, which provides high-quality, modular implementations of representative SRL techniques for humanoid robot learning. Extensive experiments on the LimX Oli humanoid robot demonstrate the effectiveness of our approach, showing significant improvements over baseline methods. These results highlight the potential of SRL-empowered RL for humanoid WBC tasks, providing valuable insights for future research in data-efficient humanoid robot learning.

Still, there are currently remaining limitations to this work. While we have demonstrated the efficacy of the PvP framework using several SRL methods, future research could explore integrating additional SRL techniques to further enhance policy learning. Additionally, recent advancements in perception-based humanoid research have shown the potential of incorporating multimodal data, such as RGB or depth images, into policy learning. We aim to extend our work to these settings, thereby expanding the capabilities of humanoid robots in more complex environments.

\clearpage
\newpage

\section*{Acknowledgments}
This work is supported, in part, by the Hong Kong SAR Research Grants Council under Grant No. PolyU 15224823, the Guangdong Basic and Applied Basic Research Foundation under Grant No. 2024A1515011524, the NSFC under Grant No. 62302246, the ZJNSFC under Grant No. LQ23F010008, the Ningbo under Grants No. 2023Z237 \& 2023CX050011 \& 2024Z284 \& 2024Z289 \& 2025Z038 \& 2025Z059, and the Ningbo Key Laboratory of Spatial Intelligence and Digital Derivative, Ningbo Institute of Digital Twin (IDT) under Grant No. S203.2.01.32.002. This work is also supported by the High Performance Computing Center at the IDT and the Eastern Institute of Technology, Ningbo. Finally, we would like to extend our sincere gratitude to Mr. Xiuyong Yao (LimX Dynamics, Shenzhen) for the enlightening discussions and valuable suggestions throughout this research. 

{
    \small
    \bibliographystyle{ieeenat_fullname}
    \bibliography{main}

\begin{thebibliography}{54}
\providecommand{\natexlab}[1]{#1}
\providecommand{\url}[1]{\texttt{#1}}
\expandafter\ifx\csname urlstyle\endcsname\relax
  \providecommand{\doi}[1]{doi: #1}\else
  \providecommand{\doi}{doi: \begingroup \urlstyle{rm}\Url}\fi

\bibitem[Chen and He(2021)]{Chen2021Simsiam}
Xinlei Chen and Kaiming He.
\newblock Exploring simple siamese representation learning.
\newblock In \emph{2021 IEEE/CVF Conference on Computer Vision and Pattern Recognition (CVPR)}, pages 15745--15753, 2021.

\bibitem[Cheng et~al.(2024)Cheng, Ji, Chen, Yang, Yang, and Wang]{cheng2024expressive}
Xuxin Cheng, Yandong Ji, Junming Chen, Ruihan Yang, Ge Yang, and Xiaolong Wang.
\newblock Expressive whole-body control for humanoid robots.
\newblock In \emph{Robotics: Science and Systems}, 2024.

\bibitem[Darvish et~al.(2023)Darvish, Penco, Ramos, Cisneros, Pratt, Yoshida, Ivaldi, and Pucci]{darvish2023teleoperation}
Kourosh Darvish, Luigi Penco, Joao Ramos, Rafael Cisneros, Jerry Pratt, Eiichi Yoshida, Serena Ivaldi, and Daniele Pucci.
\newblock Teleoperation of humanoid robots: A survey.
\newblock \emph{IEEE Transactions on Robotics}, 39\penalty0 (3):\penalty0 1706--1727, 2023.

\bibitem[Dunion et~al.(2023{\natexlab{a}})Dunion, McInroe, Luck, Hanna, and Albrecht]{dunion2023conditional}
Mhairi Dunion, Trevor McInroe, Kevin Luck, Josiah Hanna, and Stefano Albrecht.
\newblock Conditional mutual information for disentangled representations in reinforcement learning.
\newblock \emph{Advances in Neural Information Processing Systems}, 36:\penalty0 80111--80129, 2023{\natexlab{a}}.

\bibitem[Dunion et~al.(2023{\natexlab{b}})Dunion, McInroe, Luck, Hanna, and Albrecht]{dunion2023temporal}
Mhairi Dunion, Trevor McInroe, Kevin~Sebastian Luck, Josiah~P. Hanna, and Stefano~V Albrecht.
\newblock Temporal disentanglement of representations for improved generalisation in reinforcement learning.
\newblock In \emph{The Eleventh International Conference on Learning Representations}, 2023{\natexlab{b}}.

\bibitem[Echchahed and Castro(2025)]{echchahed2025a}
Ayoub Echchahed and Pablo~Samuel Castro.
\newblock A survey of state representation learning for deep reinforcement learning.
\newblock \emph{Transactions on Machine Learning Research}, 2025.
\newblock Survey Certification.

\bibitem[Higgins et~al.(2017)Higgins, Matthey, Pal, Burgess, Glorot, Botvinick, Mohamed, and Lerchner]{higgins2017beta}
Irina Higgins, Loic Matthey, Arka Pal, Christopher Burgess, Xavier Glorot, Matthew Botvinick, Shakir Mohamed, and Alexander Lerchner.
\newblock beta-vae: Learning basic visual concepts with a constrained variational framework.
\newblock In \emph{International conference on learning representations}, 2017.

\bibitem[Huang et~al.(2025)Huang, Ren, Wang, Wang, Ben, Wen, Chen, Li, and Pang]{huang2025learning}
Tao Huang, Junli Ren, Huayi Wang, Zirui Wang, Qingwei Ben, Muning Wen, Xiao Chen, Jianan Li, and Jiangmiao Pang.
\newblock Learning humanoid standing-up control across diverse postures.
\newblock In \emph{Robotics: Science and Systems}, 2025.

\bibitem[Islam et~al.(2023)Islam, Tomar, Lamb, Efroni, Zang, Didolkar, Misra, Li, Van~Seijen, Des~Combes, et~al.]{islam2023principled}
Riashat Islam, Manan Tomar, Alex Lamb, Yonathan Efroni, Hongyu Zang, Aniket~Rajiv Didolkar, Dipendra Misra, Xin Li, Harm Van~Seijen, Remi~Tachet Des~Combes, et~al.
\newblock Principled offline rl in the presence of rich exogenous information.
\newblock In \emph{International Conference on Machine Learning}, pages 14390--14421. PMLR, 2023.

\bibitem[Kingma and Welling(2014)]{kingma2014auto}
Diederik~P Kingma and Max Welling.
\newblock Auto-encoding variational bayes.
\newblock In \emph{International Conference on Learning Representations}, 2014.

\bibitem[Kuindersma et~al.(2016)Kuindersma, Deits, Fallon, Valenzuela, Dai, Permenter, Koolen, Marion, and Tedrake]{kuindersma2016optimization}
Scott Kuindersma, Robin Deits, Maurice Fallon, Andr{\'e}s Valenzuela, Hongkai Dai, Frank Permenter, Twan Koolen, Pat Marion, and Russ Tedrake.
\newblock Optimization-based locomotion planning, estimation, and control design for the atlas humanoid robot.
\newblock \emph{Autonomous robots}, 40\penalty0 (3):\penalty0 429--455, 2016.

\bibitem[Kumar et~al.(2021)Kumar, Fu, Pathak, and Malik]{kumar2021rma}
Ashish Kumar, Zipeng Fu, Deepak Pathak, and Jitendra Malik.
\newblock Rma: Rapid motor adaptation for legged robots.
\newblock 2021.

\bibitem[Lamb et~al.(2025)Lamb, Islam, Efroni, Didolkar, Misra, Foster, Molu, Chari, Krishnamurthy, and Langford]{lamb2025guaranteed}
Alex Lamb, Riashat Islam, Yonathan Efroni, Aniket~Rajiv Didolkar, Dipendra Misra, Dylan~J Foster, Lekan~P Molu, Rajan Chari, Akshay Krishnamurthy, and John Langford.
\newblock Guaranteed discovery of control-endogenous latent states with multi-step inverse models.
\newblock \emph{Transactions on Machine Learning Research}, 2025.

\bibitem[Laskin et~al.(2020)Laskin, Srinivas, and Abbeel]{laskin2020curl}
Michael Laskin, Aravind Srinivas, and Pieter Abbeel.
\newblock Curl: Contrastive unsupervised representations for reinforcement learning.
\newblock In \emph{International conference on machine learning}, pages 5639--5650. PMLR, 2020.

\bibitem[Laskin et~al.(2022{\natexlab{a}})Laskin, Liu, Peng, Yarats, Rajeswaran, and Abbeel]{laskin2022cic}
Michael Laskin, Hao Liu, Xue~Bin Peng, Denis Yarats, Aravind Rajeswaran, and Pieter Abbeel.
\newblock Unsupervised reinforcement learning with contrastive intrinsic control.
\newblock In \emph{Advances in Neural Information Processing Systems}, pages 34478--34491, 2022{\natexlab{a}}.

\bibitem[Laskin et~al.(2022{\natexlab{b}})Laskin, Liu, Peng, Yarats, Rajeswaran, and Abbeel]{laskin2022unsupervised}
Michael Laskin, Hao Liu, Xue~Bin Peng, Denis Yarats, Aravind Rajeswaran, and Pieter Abbeel.
\newblock Unsupervised reinforcement learning with contrastive intrinsic control.
\newblock \emph{Advances in Neural Information Processing Systems}, 35:\penalty0 34478--34491, 2022{\natexlab{b}}.

\bibitem[Lee et~al.(2020)Lee, Hwangbo, Wellhausen, Koltun, and Hutter]{lee2020learning}
Joonho Lee, Jemin Hwangbo, Lorenz Wellhausen, Vladlen Koltun, and Marco Hutter.
\newblock Learning quadrupedal locomotion over challenging terrain.
\newblock \emph{Science robotics}, 5\penalty0 (47):\penalty0 eabc5986, 2020.

\bibitem[Li et~al.(2025)Li, Lin, Cui, Liu, Liang, Zhu, and Huang]{li2025clone}
Yixuan Li, Yutang Lin, Jieming Cui, Tengyu Liu, Wei Liang, Yixin Zhu, and Siyuan Huang.
\newblock Clone: Closed-loop whole-body humanoid teleoperation for long-horizon tasks.
\newblock \emph{arXiv preprint arXiv:2506.08931}, 2025.

\bibitem[Liao et~al.(2025)Liao, Truong, Huang, Tevet, Sreenath, and Liu]{liao2025beyondmimic}
Qiayuan Liao, Takara~E Truong, Xiaoyu Huang, Guy Tevet, Koushil Sreenath, and C~Karen Liu.
\newblock Beyondmimic: From motion tracking to versatile humanoid control via guided diffusion.
\newblock \emph{arXiv preprint arXiv:2508.08241}, 2025.

\bibitem[Long et~al.(2024)Long, Ren, Shi, Wang, Huang, Luo, and Pang]{long2024learning}
Junfeng Long, Junli Ren, Moji Shi, Zirui Wang, Tao Huang, Ping Luo, and Jiangmiao Pang.
\newblock Learning humanoid locomotion with perceptive internal model.
\newblock \emph{arXiv preprint arXiv:2411.14386}, 2024.

\bibitem[Makoviychuk et~al.(2021)Makoviychuk, Wawrzyniak, Guo, Lu, Storey, Macklin, Hoeller, Rudin, Allshire, Handa, and State]{makoviychuk2021isaac}
Viktor Makoviychuk, Lukasz Wawrzyniak, Yunrong Guo, Michelle Lu, Kier Storey, Miles Macklin, David Hoeller, Nikita Rudin, Arthur Allshire, Ankur Handa, and Gavriel State.
\newblock Isaac gym: High performance {GPU} based physics simulation for robot learning.
\newblock In \emph{Thirty-fifth Conference on Neural Information Processing Systems Datasets and Benchmarks Track}, 2021.

\bibitem[McInroe et~al.(2025)McInroe, Sch{\"a}fer, and Albrecht]{mcinroe2025multi}
Trevor McInroe, Lukas Sch{\"a}fer, and Stefano~V Albrecht.
\newblock Multi-horizon representations with hierarchical forward models for reinforcement learning.
\newblock 2025.

\bibitem[Mittal et~al.(2023)Mittal, Yu, Yu, Liu, Rudin, Hoeller, Yuan, Singh, Guo, Mazhar, Mandlekar, Babich, State, Hutter, and Garg]{Mayank2023Orbit}
Mayank Mittal, Calvin Yu, Qinxi Yu, Jingzhou Liu, Nikita Rudin, David Hoeller, Jia~Lin Yuan, Ritvik Singh, Yunrong Guo, Hammad Mazhar, Ajay Mandlekar, Buck Babich, Gavriel State, Marco Hutter, and Animesh Garg.
\newblock Orbit: A unified simulation framework for interactive robot learning environments.
\newblock \emph{IEEE Robotics and Automation Letters}, 8\penalty0 (6):\penalty0 3740--3747, 2023.

\bibitem[Mittal et~al.(2024)Mittal, Rudin, Klemm, Allshire, and Hutter]{mittal2024symmetry}
Mayank Mittal, Nikita Rudin, Victor Klemm, Arthur Allshire, and Marco Hutter.
\newblock Symmetry considerations for learning task symmetric robot policies.
\newblock In \emph{2024 IEEE International Conference on Robotics and Automation (ICRA)}, pages 7433--7439. IEEE, 2024.

\bibitem[NVIDIA(2025)]{nvidia2025isaaclab}
NVIDIA.
\newblock Isaac lab: A gpu-accelerated simulation framework for multi-modal robot learning, 2025.

\bibitem[Oord et~al.(2018)Oord, Li, and Vinyals]{oord2018representation}
Aaron van~den Oord, Yazhe Li, and Oriol Vinyals.
\newblock Representation learning with contrastive predictive coding.
\newblock \emph{arXiv preprint arXiv:1807.03748}, 2018.

\bibitem[Pathak et~al.(2017)Pathak, Agrawal, Efros, and Darrell]{pathak2017curiosity}
Deepak Pathak, Pulkit Agrawal, Alexei~A Efros, and Trevor Darrell.
\newblock Curiosity-driven exploration by self-supervised prediction.
\newblock In \emph{International conference on machine learning}, pages 2778--2787. PMLR, 2017.

\bibitem[Romualdi et~al.(2022)Romualdi, Dafarra, L'Erario, Sorrentino, Traversaro, and Pucci]{romualdi2022online}
Giulio Romualdi, Stefano Dafarra, Giuseppe L'Erario, Ines Sorrentino, Silvio Traversaro, and Daniele Pucci.
\newblock Online non-linear centroidal mpc for humanoid robot locomotion with step adjustment.
\newblock In \emph{2022 International Conference on Robotics and Automation (ICRA)}, pages 10412--10419. IEEE, 2022.

\bibitem[Schulman et~al.(2015)Schulman, Moritz, Levine, Jordan, and Abbeel]{schulman2015high}
John Schulman, Philipp Moritz, Sergey Levine, Michael Jordan, and Pieter Abbeel.
\newblock High-dimensional continuous control using generalized advantage estimation.
\newblock \emph{arXiv preprint arXiv:1506.02438}, 2015.

\bibitem[Schulman et~al.(2017)Schulman, Wolski, Dhariwal, Radford, and Klimov]{schulman2017proximal}
John Schulman, Filip Wolski, Prafulla Dhariwal, Alec Radford, and Oleg Klimov.
\newblock Proximal policy optimization algorithms.
\newblock \emph{arXiv preprint arXiv:1707.06347}, 2017.

\bibitem[Schwarke et~al.(2023)Schwarke, Klemm, Boon, Bjelonic, and Hutter]{schwarke2023curiosity}
Clemens Schwarke, Victor Klemm, Matthijs van~der Boon, Marko Bjelonic, and Marco Hutter.
\newblock Curiosity-driven learning of joint locomotion and manipulation tasks.
\newblock In \emph{Proceedings of The 7th Conference on Robot Learning}, pages 2594--2610. PMLR, 2023.

\bibitem[Schwarke et~al.(2025)Schwarke, Mittal, Rudin, Hoeller, and Hutter]{schwarke2025rslrl}
Clemens Schwarke, Mayank Mittal, Nikita Rudin, David Hoeller, and Marco Hutter.
\newblock Rsl-rl: A learning library for robotics research.
\newblock \emph{arXiv preprint arXiv:2509.10771}, 2025.

\bibitem[Schwarzer et~al.(2021)Schwarzer, Anand, Goel, Hjelm, Courville, and Bachman]{schwarzer2021dataefficient}
Max Schwarzer, Ankesh Anand, Rishab Goel, R~Devon Hjelm, Aaron Courville, and Philip Bachman.
\newblock Data-efficient reinforcement learning with self-predictive representations.
\newblock In \emph{International Conference on Learning Representations}, 2021.

\bibitem[Sentis and Khatib(2005)]{sentis2005synthesis}
Luis Sentis and Oussama Khatib.
\newblock Synthesis of whole-body behaviors through hierarchical control of behavioral primitives.
\newblock \emph{International Journal of Humanoid Robotics}, 2\penalty0 (04):\penalty0 505--518, 2005.

\bibitem[Sentis and Khatib(2006)]{sentis2006whole}
Luis Sentis and Oussama Khatib.
\newblock A whole-body control framework for humanoids operating in human environments.
\newblock In \emph{IEEE International Conference on Robotics and Automation}, pages 2641--2648. IEEE, 2006.

\bibitem[Siekmann et~al.(2021)Siekmann, Godse, Fern, and Hurst]{siekmann2021sim}
Jonah Siekmann, Yesh Godse, Alan Fern, and Jonathan Hurst.
\newblock Sim-to-real learning of all common bipedal gaits via periodic reward composition.
\newblock In \emph{2021 IEEE International Conference on Robotics and Automation (ICRA)}, pages 7309--7315. IEEE, 2021.

\bibitem[Spaan(2012)]{spaan2012partially}
Matthijs~TJ Spaan.
\newblock Partially observable markov decision processes.
\newblock In \emph{Reinforcement learning: State-of-the-art}, pages 387--414. Springer, 2012.

\bibitem[Stooke et~al.(2021)Stooke, Lee, Abbeel, and Laskin]{stooke2021decoupling}
Adam Stooke, Kimin Lee, Pieter Abbeel, and Michael Laskin.
\newblock Decoupling representation learning from reinforcement learning.
\newblock In \emph{International conference on machine learning}, pages 9870--9879. PMLR, 2021.

\bibitem[Sun et~al.(2025{\natexlab{a}})Sun, Cao, Chen, Su, Liu, Xie, and Liu]{sun2025learning}
Wandong Sun, Baoshi Cao, Long Chen, Yongbo Su, Yang Liu, Zongwu Xie, and Hong Liu.
\newblock Learning perceptive humanoid locomotion over challenging terrain.
\newblock \emph{arXiv preprint arXiv:2503.00692}, 2025{\natexlab{a}}.

\bibitem[Sun et~al.(2025{\natexlab{b}})Sun, Chen, Su, Cao, Liu, and Xie]{sun2025learning_world}
Wandong Sun, Long Chen, Yongbo Su, Baoshi Cao, Yang Liu, and Zongwu Xie.
\newblock Learning humanoid locomotion with world model reconstruction.
\newblock \emph{arXiv preprint arXiv:2502.16230}, 2025{\natexlab{b}}.

\bibitem[Team()]{mujoco_mjx}
MuJoCo Team.
\newblock Mujoco xla.

\bibitem[Todorov et~al.(2012)Todorov, Erez, and Tassa]{todorov2012mujoco}
Emanuel Todorov, Tom Erez, and Yuval Tassa.
\newblock Mujoco: A physics engine for model-based control.
\newblock In \emph{2012 IEEE/RSJ international conference on intelligent robots and systems}, pages 5026--5033. IEEE, 2012.

\bibitem[Tong et~al.(2024)Tong, Liu, and Zhang]{tong2024advancements}
Yuchuang Tong, Haotian Liu, and Zhengtao Zhang.
\newblock Advancements in humanoid robots: A comprehensive review and future prospects.
\newblock \emph{IEEE/CAA Journal of Automatica Sinica}, 11\penalty0 (2):\penalty0 301--328, 2024.

\bibitem[Valenzuela et~al.(2024)Valenzuela, Roxas, and Wong]{Valenzuela2024EmbodyingIH}
Kirsten~Lynx Valenzuela, Samantha~Isabel Roxas, and Yung-Hao Wong.
\newblock Embodying intelligence: Humanoid robot advancements and future prospects.
\newblock In \emph{Interacci{\'o}n}, 2024.

\bibitem[Wang et~al.(2024)Wang, Luo, Zhang, and Chen]{wang2024cts}
Hongxi Wang, Haoxiang Luo, Wei Zhang, and Hua Chen.
\newblock Cts: Concurrent teacher-student reinforcement learning for legged locomotion.
\newblock \emph{IEEE Robotics and Automation Letters}, 2024.

\bibitem[Xue et~al.(2025)Xue, Dong, Liuˆ, Zhang, and Pang]{xue2025unified}
Yufei Xue, Wentao Dong, Minghuan Liuˆ, Weinan Zhang, and Jiangmiao Pang.
\newblock A unified and general humanoid whole-body controller for versatile locomotion.
\newblock In \emph{Robotics: Science and Systems}, 2025.

\bibitem[Yu et~al.(2021)Yu, Lan, Zeng, Feng, Zhang, and Chen]{yu2021playvirtual}
Tao Yu, Cuiling Lan, Wenjun Zeng, Mingxiao Feng, Zhizheng Zhang, and Zhibo Chen.
\newblock Playvirtual: Augmenting cycle-consistent virtual trajectories for reinforcement learning.
\newblock \emph{Advances in Neural Information Processing Systems}, 34:\penalty0 5276--5289, 2021.

\bibitem[Yu et~al.(2022)Yu, Zhang, Lan, Lu, and Chen]{yu2022mask}
Tao Yu, Zhizheng Zhang, Cuiling Lan, Yan Lu, and Zhibo Chen.
\newblock Mask-based latent reconstruction for reinforcement learning.
\newblock \emph{Advances in Neural Information Processing Systems}, 35:\penalty0 25117--25131, 2022.

\bibitem[Yuan et~al.(2025{\natexlab{a}})Yuan, Castanyer, Li, Jin, Zeng, and Berseth]{yuan2025rlexplore}
Mingqi Yuan, Roger~Creus Castanyer, Bo Li, Xin Jin, Wenjun Zeng, and Glen Berseth.
\newblock Rlexplore: Accelerating research in intrinsically-motivated reinforcement learning.
\newblock \emph{Transactions on Machine Learning Research}, 2025{\natexlab{a}}.

\bibitem[Yuan et~al.(2025{\natexlab{b}})Yuan, Yu, Ge, Yao, Wang, Chen, Jin, Li, Chen, et~al.]{yuan2025behavior}
Mingqi Yuan, Tao Yu, Wenqi Ge, Xiuyong Yao, Huijiang Wang, Jiayu Chen, Xin Jin, Bo Li, Hua Chen, et~al.
\newblock Behavior foundation model: Towards next-generation whole-body control system of humanoid robots.
\newblock \emph{arXiv preprint arXiv:2506.20487}, 2025{\natexlab{b}}.

\bibitem[Ze et~al.(2025)Ze, Chen, Araujo, ang Cao, Peng, Wu, and Liu]{ze2025twist}
Yanjie Ze, Zixuan Chen, Joao~Pedro Araujo, Zi ang Cao, Xue~Bin Peng, Jiajun Wu, and Karen Liu.
\newblock {TWIST}: Teleoperated whole-body imitation system.
\newblock In \emph{9th Annual Conference on Robot Learning}, 2025.

\bibitem[Zhang et~al.(2025)Zhang, Guo, Chen, Wang, Lin, Lian, Xue, Wang, Liu, Liu, et~al.]{zhang2025track}
Zhikai Zhang, Jun Guo, Chao Chen, Jilong Wang, Chenghuai Lin, Yunrui Lian, Han Xue, Zhenrong Wang, Maoqi Liu, Huaping Liu, et~al.
\newblock Track any motions under any disturbances.
\newblock \emph{arXiv preprint arXiv:2509.13833}, 2025.

\bibitem[Zheng et~al.(2024)Zheng, Salakhutdinov, and Eysenbach]{zheng2024contrastive}
Chongyi Zheng, Ruslan Salakhutdinov, and Benjamin Eysenbach.
\newblock Contrastive difference predictive coding.
\newblock In \emph{International Conference on Learning Representations}, 2024.

\bibitem[Zheng et~al.(2025)Zheng, Cheng, Liu, Li, Li, Ye, and Liu]{zheng2025transformer}
Han Zheng, Yi Cheng, Hang Liu, Jiayi Li, Yizhe Li, Linqi Ye, and Houde Liu.
\newblock Transformer-based world interaction modeling for humanoid locomotion control.
\newblock In \emph{2025 IEEE 21st International Conference on Automation Science and Engineering (CASE)}, pages 2766--2773. IEEE, 2025.

\end{thebibliography}
}


\clearpage
\appendix

\section{Algorithmic Baselines}\label{appendix:baselines}

\subsection{PPO}
Proximal policy optimization (PPO) \cite{schulman2017proximal} is an on-policy algorithm that is designed to improve the stability and sample efficiency of policy gradient methods, which uses a clipped surrogate objective function to avoid large policy updates. 

The policy loss is defined as:
\begin{equation}
\begin{aligned}
    L_{\pi}(\bm{\theta}) &= -\mathbb{E}_{\tau\sim\pi} \left[ \min\left( \rho_{t}(\bm{\theta})A_{t}, \right. \right. \\ 
    & \left. \left. {\rm clip}\left(\rho_{t}(\bm{\theta}),1-\epsilon,1+\epsilon\right)A_{t} \right) \right],
\end{aligned}
\end{equation}
where 
\begin{equation}
    \rho_{t}(\bm{\theta})=\frac{\pi_{\bm\theta}(\bm{a}_{t}|\bm{s}_{t})}{\pi_{\bm\theta_{\rm old}}(\bm{a}_{t}|\bm{s}_{t})},
\end{equation}
and $\epsilon$ is a clipping range coefficient.

Meanwhile, the value network is trained to minimize the error between the predicted return and a target of discounted returns computed with generalized advantage estimation (GAE) \cite{schulman2015high}:
\begin{equation}
    L_{V}(\bm{\phi})=\mathbb{E}_{\tau\sim\pi}\left[\left(V_{\bm\phi}(\bm{s})-V_{t}^{\rm target}\right)^{2}\right].
\end{equation}

\subsection{VAE}
Variational autoencoders (VAE) \cite{kingma2014auto} are reconstruction-based methods that encode observations $\bm{o}$ into latent variables $\bm{z}$ while enforcing a prior distribution, balancing reconstruction fidelity and regularization. The loss function of VAE is defined as
\begin{equation}
\begin{aligned}
L_{\rm VAE}
= &-\mathbb{E}_{q_{\bm\phi}(\bm{z}|\bm{o})}\!\left[\log p_{\bm\theta}(\bm{o}|\bm{z})\right]\\
&+ D_{\rm KL}\!\left(q_{\bm\phi}(\bm{z}| \bm{o}) \Vert p_{\bm\theta}(\bm{z})\right),
\end{aligned}
\end{equation}
where $q_{\bm\phi}(\bm{z}|\bm{o})$ is the encoder, $p_{\bm\theta}(\bm{o}|\bm{z})$ is the decoder, and $D_{\rm KL}$ is the Kullback–Leibler (KL) divergence.


\subsection{SPR}
SPR \cite{schwarzer2021dataefficient} is a dynamics modeling method that learns predictive latent representations by enforcing multi-step consistency between predicted and encoded future states. The loss function of SPR is defined as
\begin{equation}
\begin{aligned}
    L_{\rm SPR}
= 
    \sum_{k=1}^{K}&\Vert
        f_{\bm\theta}^{(k)}\!\big(\bm{z}_t, \bm{a}_{t:t+k-1}\big) \\
        &- \text{sg}\!\left(g_{\bm\phi}(\bm{o}_{t+k})\right)\Vert_2^2,
\end{aligned}
\end{equation}
where $\mathrm{sg}(\cdot)$ denotes the stop-gradient operation. Here, $f_{\bm\theta}$ is the online dynamics model, such that $\bm{z}_{t+1}=f_{\bm\theta}(\bm{z}_{t},\bm{a}_{t})$. $g_{\bm\phi}$ is the target dynamics model whose parameters are an exponential moving average (EMA) of the online dynamics model parameters.

\subsection{SimSiam}
SimSiam \cite{Chen2021Simsiam} is a simple self-supervised learning method based on a Siamese network architecture, designed to learn meaningful representations without the need for negative sample pairs, large batches, or momentum encoders. The architecture consists of two identical networks that process two augmented views of the same input image. A key feature of SimSiam is the use of a stop-gradient operation, which prevents the network from collapsing by ensuring that gradients do not propagate to one of the branches. The objective is to maximize the similarity between the representations of the two views, which is achieved using the negative cosine similarity loss function.

The loss function used in SimSiam is given by:

\begin{equation}
    L_{\rm SimSiam} = \frac{1}{2} \left[ - \frac{f_{\bm{\theta}}(\bm{x_1}) \cdot f_{\bm{\theta}}(\bm{x_2})}{\|f_{\bm{\theta}}(\bm{x_1})\|_2 \|f_{\bm{\theta}}(\bm{x_2})\|_2} \right]
\end{equation}

where \( f_{\bm\theta}(\bm{x}) \) represents the encoder network's output for the augmented view \( \bm{x} \). 

\section{Benchmark Design}\label{appendix:benchmark}

\begin{figure*}[ht!]
    \centering
    \begin{subfigure}{\linewidth}
        \centering
        \includegraphics[width=\textwidth]{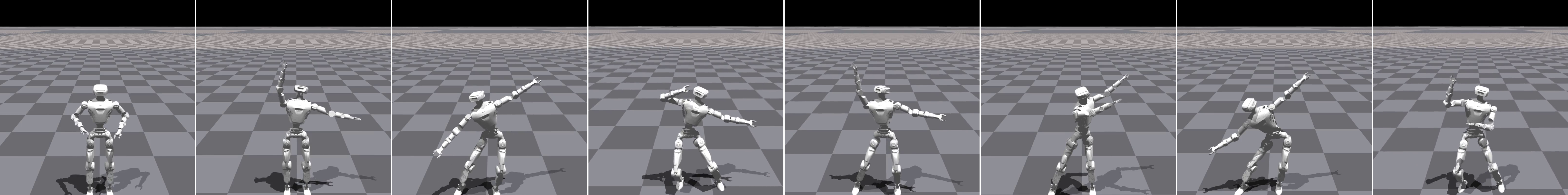}
    \end{subfigure}
    \vfill
    \begin{subfigure}{\linewidth}
        \centering
        \includegraphics[width=\textwidth]{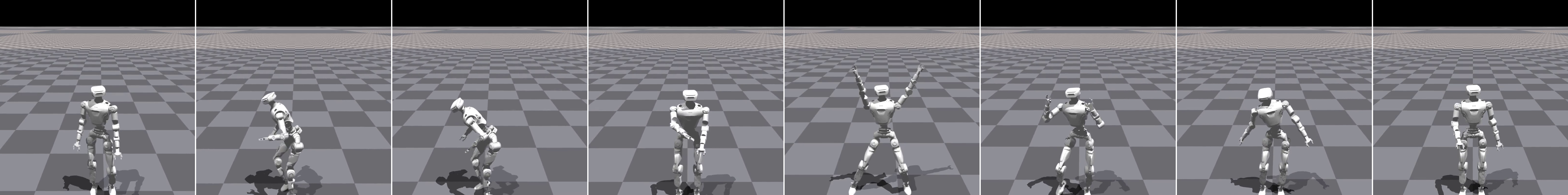}
    \end{subfigure}
    \vfill
    \begin{subfigure}{\linewidth}
        \centering
        \includegraphics[width=\textwidth]{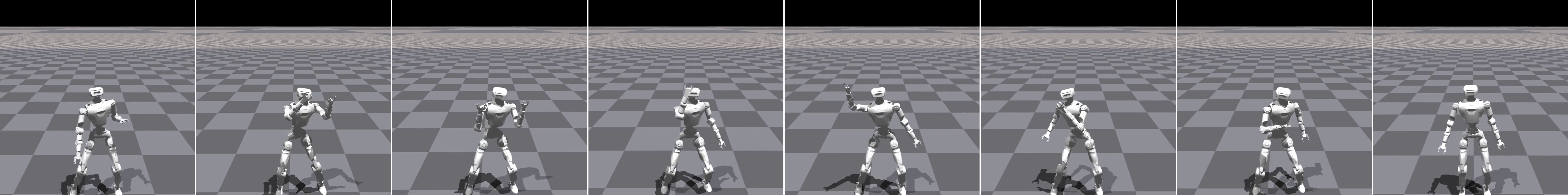}
    \end{subfigure}
    \caption{Example screenshots of the motion capture data.}
    \label{fig:mocap}
\end{figure*}

\begin{table}[h!]
\centering
\caption{Key reward terms utilized in \textit{LimX-Oli-31dof-Velocity} task.}
\label{tb:reward_terms_vel}
\small	
\begin{tabularx}{\linewidth}{XXl}
\toprule
\textbf{Term}            & \textbf{Formulation}              & \textbf{Weight} \\ \midrule
Linear velocity tracking  & $\exp\left(-\tfrac{\Vert \bm{v}_{xy} - \bm{v}^{\rm cmd}_{xy} \Vert^2}{2\sigma^2}\right)$ & 1.0 \\
Angular velocity tracking & $\exp\!\left(-\tfrac{(\omega_z-\omega_z^{\rm cmd})^2}{2\sigma^2}\right)$                 & 0.5 \\
Base height              & $(h-h^\star)^2$                   & 0.5             \\
Linear velocity $(z)$    & $\|v_z\|^2$                       & -2e-3           \\
Angular velocity $(x,y)$ & $\Vert\bm{\omega}_{xy}\Vert^2$    & -0.15           \\
Action smoothness              & $\|\bm{a}_t-2\bm{a}_{t-1}-\bm{a}_{t-2}\|^2$     & -2.5e-3         \\
Joint velocity           & $\|\dot{\bm{q}}\|^2$              & -1e-3           \\
Joint acceleration       & $\|\ddot{\bm{q}}\|^2$             & -5e-7           \\
Joint deviation          & $\sum_j |q_j - q_j^{\text{def}}|$ & -0.1            \\
Joint power              & $|\bm{\tau}||\dot{\bm q}|^{T}$    & -2.5e-7         \\
Joint torque             & $\Vert\bm{\tau}\Vert_{2}^{2}$     & -4.0e-7         \\
Joint position limits    & $\sum_j \Delta_j$                 & -0.2            \\
Joint velocity limits    & $\sum_j \dot{q}_j$                & -0.025          \\ \bottomrule
\end{tabularx}
\end{table}
Let $\bm{q}$ denote the joint positions, $\dot{\bm{q}}$ the joint velocities, $\ddot{\bm{q}}$ the joint accelerations, $\bm{\tau}$ the joint torque, $\bm{v}_{xy}$ the base linear velocity, $\bm{\omega}_{xy}$ the base angular velocity on the xy-axis, $h^*$ the expected base height, $\delta_1,\delta_2$ the roll and pitch of the waist joint, and $\Delta$ the the absolute value of the difference between the joint position and the soft limits. The following tables illustrate the reward terms and components of the proprioceptive states and the privileged states of the two tasks.

\subsection{Velocity Tracking Task}

\begin{table}[h!]
\centering
\caption{The details of the proprioceptive state and privileged state of the \textit{LimX-Oli-31dof-Velocity} task. Here, we stack 5 consecutive proprioceptive states as the input of the policy encoder to ensure robustness.}
\label{tb:task_state_vel}
\begin{tabular}{ll}
\toprule
  \textbf{Proprioceptive State} &
  \textbf{Privileged State} \\ \midrule
  \begin{tabular}[c]{@{}l@{}}base\_ang\_vel (3x5)\\ projected\_gravity (3x5)\\ velocity\_commands (3x5)\\ joint\_pos (31x5)\\ joint\_vel (31x5)\\ actions (31x5)\\ gait (5)\end{tabular} &
  \begin{tabular}[c]{@{}l@{}}base\_lin\_vel (3)\\ base\_ang\_vel (3)\\ projected\_gravity (3)\\ velocity\_commands (3)\\ joint\_pos (31)\\ joint\_vel (31)\\ actions (31)\\ gait (5)\end{tabular} \\ 
\bottomrule
\end{tabular}
\end{table}

\subsection{Motion Imitation Task}
\begin{table}[h!]
\centering
\caption{Key reward terms utilized in \textit{LimX-Oli-31dof-Mimic} task.}
\label{tb:reward_terms_mimic}
\small
\begin{tabularx}{\linewidth}{XXl}
\toprule
\textbf{Term}         & \textbf{Formulation}          & \textbf{Weight} \\ \midrule
Position tracking                & $\exp\left(-\tfrac{\Vert \bm{q} - \bm{q}^{\rm ref} \Vert^2}{2\sigma^2}\right)$       & 2.0 \\
Feet distance tracking           & $\exp\left(-\tfrac{\vert d - d^{\rm ref} \vert^2}{\sigma}\right)$       & 0.5 \\
Waist pitch orientation tracking & $\exp\left(-\sum_{i=1}^{2}|\delta_{i}-\delta_{i}^{\rm ref}|\right)$ & 0.5 \\
Action rate           & $\|\bm{a}_t-\bm{a}_{t-1}\|^2$ & -0.001          \\
Joint velocity        & $\|\dot{\bm{q}}\|^2$          & -0.5e-3         \\
Joint acceleration    & $\|\ddot{\bm{q}}\|^2$         & -1.0e-7         \\
Joint Torque          & $\Vert\bm{\tau}\Vert_{2}^{2}$ & -1.0e-5         \\
Joint position limits & $\sum_j \Delta_j$             & -1.0            \\
Joint torque limits   & $\sum_j \tau_j$               & -0.01           \\
Joint velocity limits & $\sum_j \dot{q}_j$            & -0.2            \\ \bottomrule
\end{tabularx}
\end{table}

\begin{table}[h!]
\centering
\caption{The details of the proprioceptive state and privileged state of the \textit{LimX-Oli-31dof-Mimic} task.}
\label{tb:task_state_mimic}
\begin{tabular}{ll}
\toprule
\textbf{Proprioceptive State} &
  \textbf{Privileged State} \\ \midrule
\begin{tabular}[c]{@{}l@{}}base\_ang\_vel (3)\\ projected\_gravity (3)\\ joint\_pos (31)\\ joint\_vel (31)\\ actions (31)\\ mimic reference (69)\end{tabular} &
  \begin{tabular}[c]{@{}l@{}}base\_lin\_vel (3)\\ base\_ang\_vel (3)\\ base\_pos\_z (1)\\ body\_mass (40)\\ base\_quat (6)\\ projected\_gravity (3)\\ velocity\_commands (3)\\ joint\_pos (31)\\ joint\_vel (31)\\ actions (31)\\ previous actions (31)\\ mimic reference (69)\end{tabular} \\ \bottomrule
\end{tabular}
\end{table}

\section{Experimental Setup}\label{appendix:setup}

\subsection{PPO}
PPO \cite{schulman2017proximal} is selected as the backbone RL algorithm for all the SRL methods. Table~\ref{tb:ppo_arch} illustrates the network architectures of the policy network and value network, and Table~\ref{tb:ppo_hp} lists the hyperparameters used for the two humanoid WBC tasks. Notably, these configurations remain fixed for all the experiments to isolate the effects of SRL methods.

\begin{table}[h!]
\centering
\caption{The architectures of the policy and value network, which remain fixed for all the experiments. Here, "O. D." represents "On-demand".}
\label{tb:ppo_arch}
\begin{tabular}{lll}
\toprule
\textbf{Part} & \textbf{Policy Network} & \textbf{Value Network} \\ \midrule
              & Linear(O. D., 512)        & Linear(O. D., 512)       \\
              & ELU()                   & ELU()                  \\
Encoder       & Linear(512, 256)        & Linear(512, 256)       \\
              & ELU()                   & ELU()                  \\
              & Linear(256, 128)        & Linear(256, 128)       \\ \midrule
              & Linear(128, 128)        & Linear(256, 128)       \\
Head          & ELU()                   & ELU()                  \\
              & Linear(128, 31)         & Linear(128, 1)         \\ \bottomrule
\end{tabular}
\end{table}

\begin{table}[h!]
\centering
\caption{The PPO hyperparameters for the two tasks, which remain fixed for all experiments.}
\label{tb:ppo_hp}
\begin{tabular}{ll}
\toprule
\textbf{Hyperparameter}    & \textbf{Value} \\ \midrule
Reward normalization       & Yes            \\
LSTM                       & No             \\
Maximum Episodes           & 30000          \\
Episode steps              & 32             \\
Number of workers          & 1              \\
Environments per worker    & 4096           \\
Optimizer                  & Adam           \\
Learning rate              & 1e-3           \\
Learning rate scheduler    & Adaptive       \\
GAE coefficient            & 0.95           \\
Action entropy coefficient & 0.01           \\
Value loss coefficient     & 1.0            \\
Value clip range           & 0.2            \\
Max gradient norm          & 0.5            \\
Number of mini-batches     & 4              \\
Number of learning epochs  & 5              \\
Desired KL divergence      & 0.01           \\
Discount factor            & 0.99           \\ \bottomrule
\end{tabular}
\end{table}

\subsection{PPO+PvP}
For PvP, we utilize the root linear velocity relative to the world coordinate system as privileged information for contrastive learning, while the root orientation information is also involved in the motion imitation task. Accordingly, we attach the zero mask to the proprioceptive state in the whole training to align its dimension with the privileged state. For the loss coefficient, we run an initial hyperparameter search over $\{0.1,0.5,1.0\}$ and use $0.5$ as the baseline setting.

\subsection{PPO+SimSiam}

For SimSiam \cite{Chen2021Simsiam}, we run an initial hyperparameter search over the loss coefficient $\{0.1,0.5,1.0\}$ and the data augmentation operation \{random\_masking, gaussian\_noise, random\_amplitude\_scaling, identity\_mapping\}. Then the loss coefficient of $0.5$ and $\text{random\_masking},\text{identity\_mapping}$ operation are used as the baseline settings. This is because the proprioceptive state is subjected to the domain randomization in the simulator, which can also be considered data augmentation.

\subsection{PPO+SPR}
For SPR \cite{schwarzer2021dataefficient}, we run an initial hyperparameter search over the loss coefficient $\{0.1,0.5,1.0\}$, the data augmentation operation \{random\_masking, gaussian\_noise, random\_amplitude\_scaling, identity\_mapping\}, the number of prediction steps $\{1,5,10,15\}$, and whether to use a average loss. Then the loss coefficient of $0.5$, $\text{gaussian\_noise}$ operation, and the number of prediction steps of $5$ are used as the baseline settings.

\subsection{PPO+VAE}
For VAE \cite{kingma2014auto}, we simply run an initial hyperparameter search over the loss coefficient $\{0.1,0.5,1.0\}$ and use $0.1$ as the baseline setting.



\section{Ablation Studies}\label{sec:ablation}
\subsection{Comparison with Teacher-Student Distillation}
\begin{figure}[h!]
    \centering
    \vspace{-15pt}
        \includegraphics[width=\linewidth]{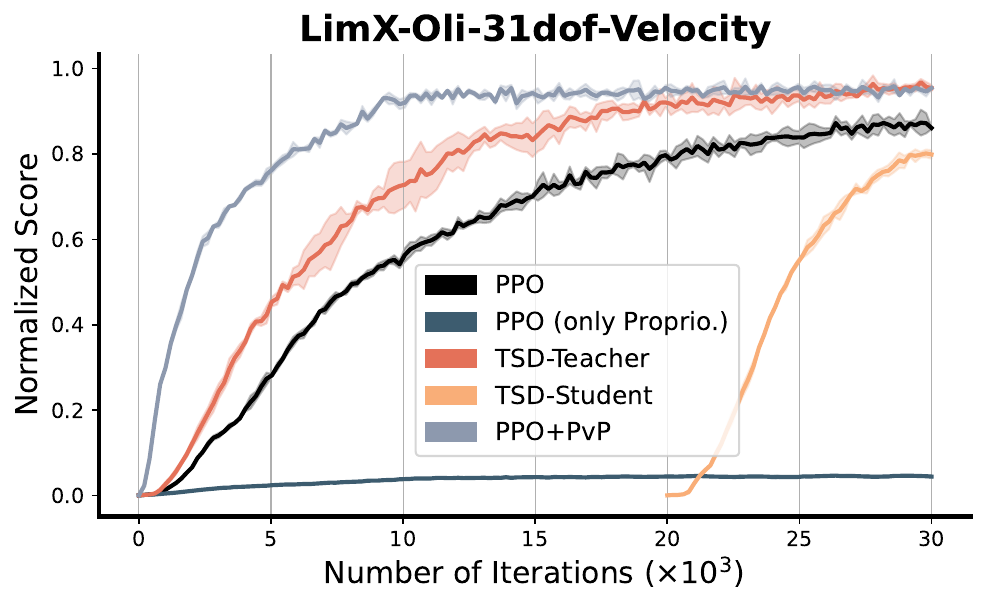}
        \vspace{-15pt}
        \caption{Training progress comparison between the teacher-student distillation method and PvP in the \textit{LimX-Oli-31dof-Velocity} task. The solid line and shaded region denote the mean and standard deviation, respectively.}
        \label{fig:retbuttal_limx_plane_tsd}
\end{figure}

Both teacher-student distillation (TSD) \cite{kumar2021rma,lee2020learning} and PvP aim to leverage privileged information to guide representation learning, thereby reducing the complexity of learning for high-dimensional humanoid control. However, TSD has several critical limitations. The student's performance ceiling is strictly constrained by the teacher's quality and inductive biases, and any sub-optimality or observation mismatches in the teacher are directly inherited by the student. As shown in the experiments below, the student fails to match the teacher's performance and exhibits a significant gap after distillation. Moreover, strict alignment objectives can over-regularize the student and suppress exploration, causing the student to collapse toward conservative or averaged behavior rather than discovering alternative (potentially better) solutions. Finally, the teacher pre-training stage is often compute-intensive, limiting reproducibility and rapid iteration as tasks scale up.

\begin{figure}[h!]
    \centering
        \vspace{-15pt}
        \includegraphics[width=\linewidth]{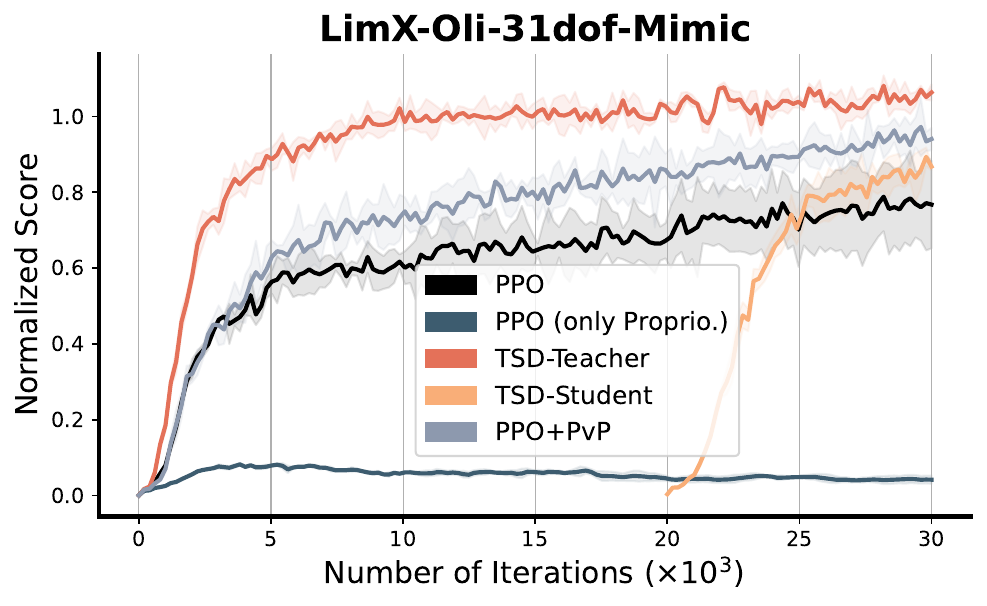}
        \vspace{-15pt}
        \caption{Training progress comparison between the teacher-student distillation method and PvP in the \textit{LimX-Oli-31dof-Mimic} task. The solid line and shaded region denote the mean and standard deviation, respectively.}
        \label{fig:retbuttal_limx_mimic}
\end{figure}

\begin{figure}[h!]
    \centering
        \includegraphics[width=\linewidth]{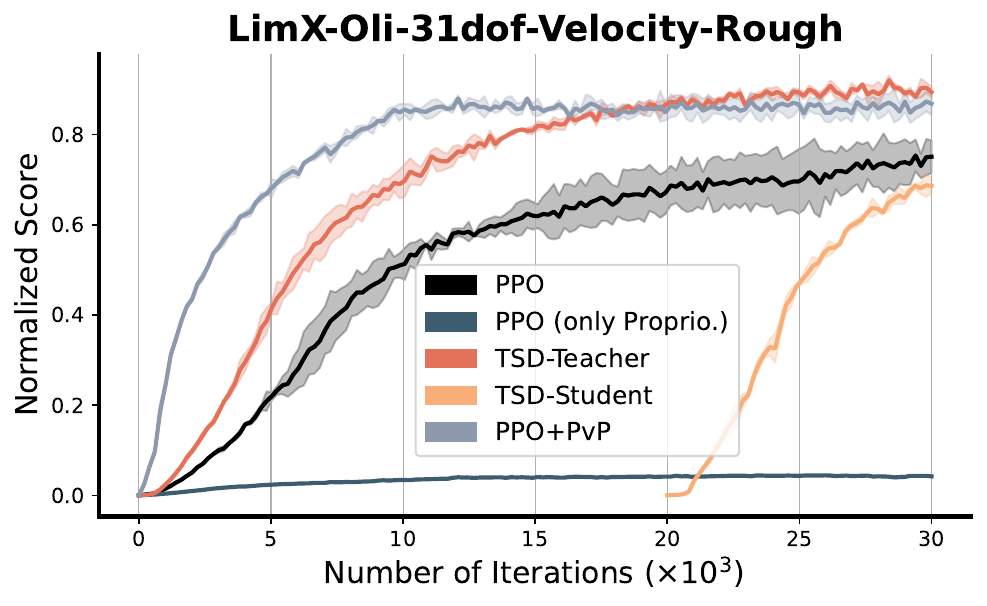}
        \caption{Training progress comparison between the teacher-student distillation method and PvP in the \textit{LimX-Oli-31dof-Velocity-Rough} task. The solid line and shaded region denote the mean and standard deviation, respectively.}
        \label{fig:retbuttal_limx_terrain_tsd}
\end{figure}

In contrast, PvP facilitates a synergy between representation learning and policy optimization. Unlike the disjointed two-stage pipeline of TSD, PvP integrates contrastive representation learning directly into the RL framework. This enables the latent space to evolve alongside the policy's exploration, rather than merely replicating a static teacher. By optimizing a contrastive objective that captures structural invariants, the representation and policy mutually enhance each other. The policy benefits from a more discriminative latent space, while the representation adaptively prioritizes task-relevant features discovered during training. 

To demonstrate PvP's advantage over the TSD method, we conduct ablation experiments on the two designed humanoid WBC tasks. As illustrated in Figure~\ref{fig:retbuttal_limx_plane_tsd} and Figure~\ref{fig:retbuttal_limx_mimic}, there is a significant performance gap between the teacher and student policies, and consistently outperforms the student policy. Notably, the PPO agent that relies solely on proprioceptive state information fails to learn in both tasks, highlighting the need for privileged information to support learning.

\subsection{Evaluation on More Diverse Tasks}
\begin{figure}[h!]
    \centering
        \includegraphics[width=\linewidth]{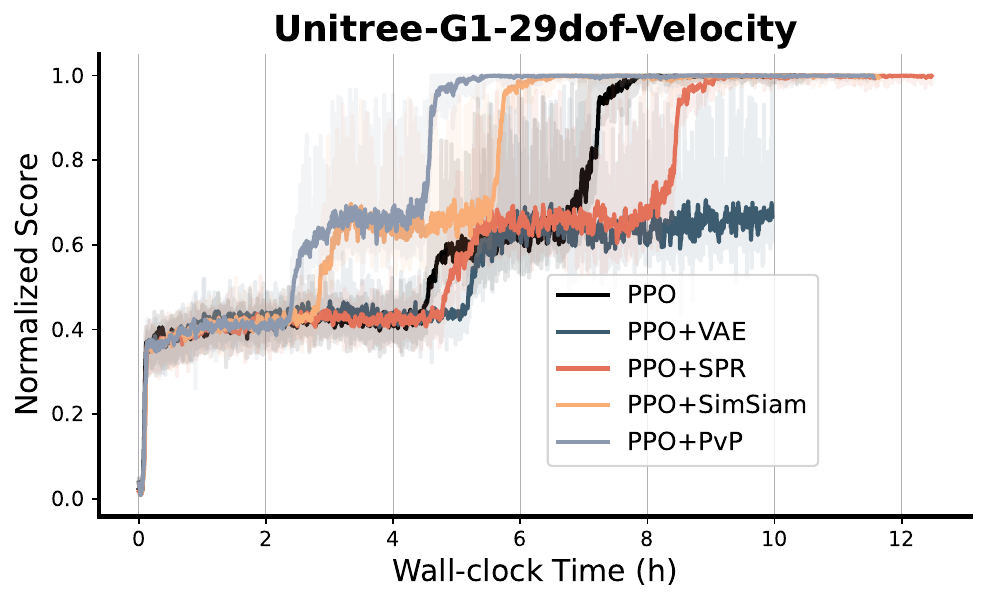}
        \vspace{-20pt}
        \caption{Training progress comparison between the vanilla PPO agent and its combination with four SRL methods on the \textit{Unitree-G1-29dof-Velocity} task. The solid line and shaded region denote the mean and standard deviation, respectively.}
        \label{fig:rebuttal_g1_time}
\end{figure}

To evaluate PvP on more diverse scenarios and robot platforms, we introduce two new tasks: \textit{LimX-Oli-31dof-Velocity-Rough} and \textit{Unitree-G1-29dof-Velocity}. The former is a variant of the previously defined velocity-tracking task that introduces rough terrain, while the latter uses a different robot platform. As illustrated in Figure~\ref{fig:retbuttal_limx_terrain_tsd} and Figure~\ref{fig:rebuttal_g1_time}, PvP excels in the two tasks in terms of final performance and time cost.

\subsection{Impact of the $\lambda$ on PvP}

\begin{figure}[h!]
    \centering
        \includegraphics[width=\linewidth]{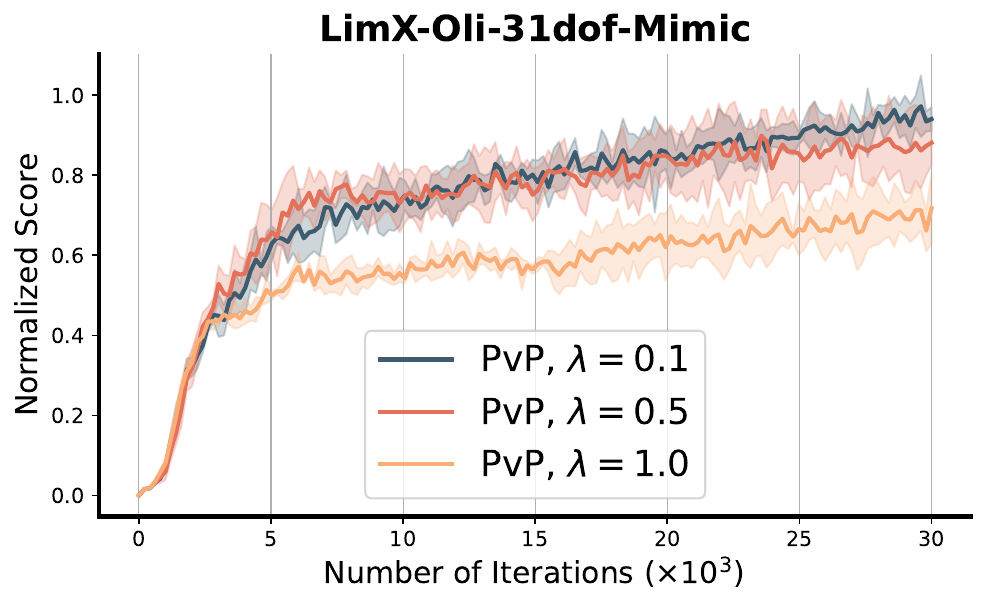}
        \vspace{-20pt}
        \caption{Performance comparison of the PvP method with different weighting coefficients on the \textit{LimX-Oli-31dof-Mimic} task. The solid line and shaded region denote the mean and standard deviation, respectively.}
        \label{fig:retbuttal_limx_mimic_lambd}
\end{figure}

Furthermore, we investigate the impact of the weighting coefficients on the performance of our PvP method. As shown in Figure~\ref{fig:retbuttal_limx_mimic_lambd}, $\lambda=0.1$ achieves the best final performance and sample efficiency, indicating that PvP benefits from a relatively smaller $\lambda$.

\end{document}